%% file: main.tex
\pdfoutput=1

\documentclass[11pt]{article}

\usepackage{acl}

\usepackage{times}
\usepackage{latexsym}
\usepackage[nointegrals]{wasysym}

\usepackage[T1]{fontenc}

\usepackage[utf8]{inputenc}

\usepackage{microtype}

\usepackage{paralist}

\usepackage{enumitem}
\usepackage{xspace}

\usepackage{booktabs} 

\usepackage{wasysym} 
\usepackage{array}
\usepackage{multirow}
\usepackage{subfig}
\usepackage{tikz}
\usepackage{pgfplots}
\usepackage{float}
\usepackage{arydshln}
\usepackage{centernot}
\usepackage{bbm}
\usepackage[normalem]{ulem}
\usepackage{tabulary}

\newcommand{\skippedDetails}[1]{}

\definecolor{orange}{rgb}{1,0.5,0}
\definecolor{mdgreen}{rgb}{0.05,0.6,0.05}
\definecolor{mdblue}{rgb}{0,0,0.7}
\definecolor{dkblue}{rgb}{0,0,0.5}
\definecolor{dkgray}{rgb}{0.3,0.3,0.3}
\definecolor{slate}{rgb}{0.25,0.25,0.4}
\definecolor{gray}{rgb}{0.5,0.5,0.5}
\definecolor{ltgray}{rgb}{0.7,0.7,0.7}
\definecolor{purple}{rgb}{0.7,0,1.0}
\definecolor{lavender}{rgb}{0.65,0.55,1.0}

\usepackage{amssymb}
\usepackage{pifont}

\usepackage{amsmath}

\newcommand{\quash}[1]{}

\usepackage[para]{footmisc}
\usepackage{arydshln}

\newcommand{\theTitle}{\datasetName: Evaluating Multilingual Inference for Indian Languages}

\newcommand{\datasetName}{{\sc IndicXNLI}\xspace}

\title{\theTitle}

\author {
    Divyanshu Aggarwal\textsuperscript{\rm 1\thanks{Equal Contribution}}~,~
    Vivek Gupta\textsuperscript{\rm {2*}\thanks{ Corresponding Author}}~,~
    Anoop Kunchukuttan\textsuperscript{\rm {3}}
    \\\textsuperscript{\rm 1}Delhi Technological University,~
     \textsuperscript{\rm 2}University of Utah,~
    \textsuperscript{\rm 3}Microsoft Research\\
    divyanshuggrwl@gmail.com; vgupta@cs.utah.edu; ankunchu@microsoft.com
}

\begin{document}
\maketitle
\begin{abstract}
While Indic NLP has made rapid advances recently in terms of the availability of corpora and pre-trained models, benchmark datasets on standard NLU tasks are limited. To this end, we introduce \datasetName, an NLI dataset for 11 Indic languages. It has been created by high-quality machine translation of the original English {\sc XNLI} dataset and our analysis attests to the quality of \datasetName. By finetuning different pre-trained LMs on this \datasetName, we analyze various cross-lingual transfer techniques with respect to the impact of the choice of language models, languages,  multi-linguality, mix-language input, etc. These experiments provide us with useful insights into the behaviour of pre-trained models for a diverse set of languages.   
\end{abstract}

\input{introduction}
\input{dataset}
\input{experiments_v2}

\input{results_analysis}

\input{error_analysis}

\input{motivation}

\input{relatedwork}

\bibliography{anthology,custom}
\bibliographystyle{acl_natbib}
\appendix
\input{appendix}

\end{document}

%% file: introduction.tex
\section{Introduction}
\label{ref:introduction}

Natural Language Inference (NLI) is a well-studied NLP task ~\citep{dagan2013recognizing} that assesses if a premise entails, negates, or is neutral towards the hypothesis statement. The task is well suited for evaluating semantic representations of state-of-the-art transformers \citep{vaswani2017attention}  models such as BERT \citep{devlin2019bert,Radford2018ImprovingLU}. Two large scale datasets, such as SNLI \citep{snli:emnlp2015} and MultiNLI \citep{N18-1101}, has recently been developed to enhanced the relevance of NLI task.

With the availability of multi-lingual pre-trained language models like mBERT \citep{devlin2019bert}, and XLM-RoBERTa \citep{conneau2020unsupervised} promising cross-lingual transfer and universal models, multi-lingual NLP has recently gained a lot of attention. However, most languages have a scarcity of datasets resources.
Some multi-lingual datasets have attempted to fill this gap, including XNLI \citep{conneau2018xnli} for NLI, XQUAD \citep{dumitrescu2021liro}, MLQA \citep{lewis-etal-2020-mlqa} for question answering, and PAWS-X for paraphrase identification \citep{yang-etal-2019-paws}.
In many practical circumstances, training sets for non-English languages are unavailable, hence cross-lingual zero-shot evaluation benchmarks such as XTREME \citep{hu2020xtreme}, XTREME-R \citep{ruder-etal-2021-xtreme}, and XGLUE \citep{liang-etal-2020-xglue} have been suggested to use these datasets.

However, NLI datasets are not available for major Indic languages. The only exceptions are the test/validation sets in the XNLI (hi and ur), TaxiXNLI (hi) \cite{k2021analyzing} and MIDAS-NLI \cite{uppal-etal-2020-two} datasets.
Furthermore, because MIDAS-NLI is based on sentiment data recasting, hypotheses are not linguistically diverse and span limited reasoning. In this work, we address this gap by introducing \datasetName, an NLI dataset for \emph{Indic} languages. \datasetName consists of English {\sc XNLI} data translated  into eleven \emph{Indic} languages. We use \datasetName to evaluate \emph{Indic}-specific models (trained only on \emph{Indic} and English languages) such as IndicBERT \cite{kakwani-etal-2020-indicnlpsuite} and MuRIL \cite{khanuja2021muril}, as well as generic (train on non-\emph{Indic} languages) such as mBERT and XLM-RoBERTa. Furthermore, we experimented with several training strategies for each multi-lingual model. Our experimental results answers multiple important questions regarding effective training for \emph{Indic} NLI. Our contributions are as follows:

\begin{itemize}
\setlength\itemsep{0.0em}
\item We introduce \datasetName, an NLI benchmark dataset for eleven prominent Indo-Aryan \emph{indic} languages from the Indo-European and Dravidian language families.
 
\item We investigate several strategies to train multi-lingual models for NLI tasks on \datasetName. We also explore models cross-lingual NLI transfer ability across \emph{Indic} languages and Intra-Bilingual NLI ability of pretrained multi-lingual language models. 
\end{itemize}

The \datasetName dataset, along with scripts, is available at \url{https://github.com/divyanshuaggarwal/indicxnli}.

%% file: dataset.tex
\section{The \datasetName dataset}
\label{sec:dataset}

We created \datasetName, a NLI data set for \emph{Indic} languages. \datasetName is similar to existing {\sc XNLI} dataset in shape/form, but focusses on \emph{Indic} language family. \datasetName include NLI data for eleven major \emph{Indic} languages that includes Assamese (`as'), Gujarat (`gu'), Kannada (`kn'), Malayalam (`ml'), Marathi (`mr'), Odia (`or'), Punjabi (`pa'), Tamil (`ta'), Telugu (`te'), Hindi (`hi'), and Bengali (`bn'). Next we describe the \datasetName construction and its validation in details.



\subsection{\datasetName Construction.} 
To create \datasetName, we follow the approach of the XNLI dataset and translate the English {\sc XNLI} dataset (premises and hypothesis) to eleven \emph{Indic}-languages. We use the IndicTrans \cite{ramesh2021samanantar}, a state-of-the-art, publicly available translation model for Indic languages, for translating from English to \emph{Indic} languages. The train (392,702), validation (2,490), and evaluation sets (5,010) of English {\sc XNLI} were translated from English into each of the eleven \emph{Indic} languages.
IndicTrans is a large Transformer-based sequence to sequence model. It is trained on Samanantar dataset \cite{ramesh2021samanantar}, which is the largest parallel multi-lingual corpus over eleven \emph{Indic} languages. IndicTrans outperforms other open-source models based on  mBART \cite{liu-etal-2020-multilingual-denoising} and mT5 \citep{xue-etal-2021-mt5} for \emph{Indic} language translations and is competitive with paid translation models such as  Google-Translate or Microsoft-Translate on several benchmarks \cite{ramesh2021samanantar}.  
Our choice of IndicTrans was motivated by \emph{cost, language coverage and speed}, refer \S \ref{sec:further-discussion}. 

\subsection{\datasetName Validation.} 
While translation may lose the semantic link between the sentences, recent study by \citet{k2021analyzing} disproved this. \citet{k2021analyzing} qualitative analysis illustrate that when a high-quality machine translation system is utilized, classification labels and reasoning categories are only minimally altered by one or two tokens for translated NLI datasets. We also demonstrate the high quality of IndicTrans translation for \datasetName in two ways \begin{inparaenum}[(a.)]\item  manual human validation and, \item automatic metric BERTScore \cite{bert-score} \end{inparaenum}. Our validation approach guarantee correctness for the \datasetName labels. Next, we'll discuss on how to evaluate IndicTrans translations.

\begin{table}[h!]
\vspace{-0.5em}
\small
\centering
\setlength{\tabcolsep}{3.5pt}
\begin{tabular}{l|rrrrrrrrrrr}\toprule
\textbf{Score} & \textbf{hi} & \textbf{te} & \textbf{pa} & \textbf{bn} & \textbf{as} & \textbf{gu} & \textbf{ta} &\textbf{ml} & \textbf{kn} & \textbf{mr}& \textbf{or}
 \\\midrule
\textbf{HS1} & 88 & 88 & 91 & 87 & 87 & 89 & 89 & 87 & 89 & 86 & 88\\
\textbf{HS2} & 81 & 84 & 93 & 83 & 84 & 89 & 87 & 87 & 87 & 87 & 90 \\
\hdashline
\textbf{PC} & 73 & 73 & 89 & 79 & 78 & 79 & 76 & 85 & 83 & 83 & 75 \\
\textbf{SC} & 82 & 87 & 94 & 90 & 88 & 85 & 88 & 93 & 86 & 89 & 85\\
\bottomrule
\end{tabular}
\caption{\small Human Validation Score ($\times$ 10$^{-2}$): \textbf{HS1}, \textbf{HS2} represents human1, human2 annotation score respectively. \textbf{PC} and \textbf{SC} represents Pearson and Spearman correlation respectively.}
\label{tab:valid_scores}
\end{table}

\paragraph{\bf Human Validation:} We followed SemEval-2016 Task-I \citep{agirre-etal-2016-semeval} guidelines. We hired 2 annotators per languages and calculated the Pearson \citep{pearson-correlation} and Spearman \citep{spearman-correlation} correlation over annotations scores of sentences.

\paragraph{\bf Diverse Sampling:} Since human validation is time-consuming and expensive. We sampled 100 diverse sentences of the test set for validation. We apply the Determinantal Point Process \cite{2012} (DPP) over sentence representations for diverse sampling. DPP maximizes coverage volume using a minimal sampled set, thus guaranteeing diversity during sampling. We first used sentence transformers to convert data to BERT embeddings, and then use k-DPP \cite{10.5555/3104482.3104632} with k = 100 to sample 100 examples. Using DPP for diverse sampling is a cost-effective method of evaluating translation quality. For scoring guidelines refer to Appendix \S\ref{sec:human-eval-guidelines-appendix}. 

\paragraph{\bf Hiring Experts:} We recruited, 2 speakers for each of the 11 \emph{indic} languages as annotators. These professional annotators are multilingual (English, \emph{Indic}) and fluent in both mother-tongue \emph{indic} and English language. The remuneration paid was 6.6 cents per sentence for each \emph{indic} language. Demographic information on annotators will be released together with the dataset. 

\paragraph{\bf Evaluation:} Table \ref{tab:valid_scores} shows the final human evaluation scores. In general, we see that average human scores is more than 0.85 for all languages. The Pearson and Spearman Correlation values are more than 0.7 and 0.8 for all languages respectively. High human ratings and high correlation between the annotations support high quality IndicTrans translation, hence validating \datasetName quality.

\begin{table}[h!]
\small
\centering
\setlength{\tabcolsep}{3.5pt}
\begin{tabular}{l|rrrrrrrrrrr}
\toprule
\textbf{BS} & \textbf{hi} & \textbf{te} & \textbf{pa} & \textbf{bn} & \textbf{as} & \textbf{gu} & \textbf{ta} &\textbf{ml} & \textbf{kn} & \textbf{mr}& \textbf{or}
 \\\midrule
\textbf{ET$^{GT}$} & 94 & 93 & 92 & 94 & NA & 94 & 94 & 94 & 94 & 94 &  94\\
\textbf{ET$^{IT}$} & 98 & 94 & 94 & 98 & 93 & 94 & 94 & 94 & 94 & 93 & 93\\
\hdashline
\textbf{ML$^{GT}$} & 90 & 88 & 86 & 89 & NA & 89 & 86 & 85 & 88 & 87 &  82 \\
\textbf{ML$^{IT}$} & 96 & 87 & 88 & 96 & 85 & 96 & 87 & 87 & 87 & 86 & 86\\

\bottomrule
\end{tabular}
\caption{\textbf{BS} represent BERTScore (F1-Score $\times$ 10$^{-2}$)  for \textbf{EngTrans} (ET) and \textbf{Multilingual} (ML) strategies. Superscript ${^{GT}}$ and ${^{IT}}$ represent Google Translate and IndicTrans models respectively.}
\label{tab:automatic_scores}
\end{table}

\paragraph{\bf Automatic Validation:} Given the absence of \emph{Indic} language {\sc XNLI} reference data, we use BERTScore similarity between the original English and English translated \datasetName for automatic evaluation. Here too, we use the IndicTrans model for translating \datasetName into English. This approach estimates the upper bound on error for the English to \emph{Indic} translation (i.e. \datasetName quality), as it approximates the combined error of both English to \emph{Indic} translation (\datasetName creation), and \emph{Indic} to English translation (evaluation) \citep{10.5555/1667583.1667625,7433246,edunov-etal-2020-evaluation, doi:10.1080/13645579.2016.1252188}. We utilize BERTScore for assessment since it correlates better with human judgment at the sentence level than BLEU \citep{bert-score,10.3115/1073083.1073135}.

We evaluate two translation models,  Google Translate and IndicTrans  on the testsets of \datasetName dataset. We incorporate Google Translate\footnote{Google Translate was accessed on 30th June 2022} to demonstrate IndicTrans's competitiveness in comparison to commercial translation approaches. In Table \ref{tab:automatic_scores}, we used two evaluation strategies for our evaluation \begin{inparaenum}[(a.)] \item \emph{EngTrans}: which take the \datasetName sentence and translated it back to English using BERT model. \item \emph{Multilingual}:  directly compare the English sentences with multilingual \datasetName sentences using mBERT model. \end{inparaenum} 

\noindent On \emph{Indic} languages, we notice that IndicTrans is comparable to, and sometimes outperforms, Google Translate. Additionally, when results are compared in a Multilingual setting, we observe a marginal decrement in scores. This can be because mBERT does not produce as precise multilingual embedding as BERT does for English. Additionally, we see a similar pattern in the distribution of scores across languages for both assessment strategies on both models. We also computed the BERTScore (using mBERT) between the Hindi test set of {\sc XNLI} and \datasetName was found to be 0.87, supporting the high quality of \datasetName.

\paragraph{\textbf{Why Machine Translation?}}
While machine translation is an extremely convenient and quick method for creating a synthetic dataset for multi-lingual NLI tasks for low-resource languages such as indic, they are prone to contain some `translationese' errors despite being meaning-preserving translations\cite{graham-etal-2020-statistical}. However, similar mistakes are still conceivable with manual translation, since humans with knowledge of both English and Indian languages may translate the text with `translationese' tendencies owing to mother tongue impact \citep{IJET23926}. 
The ideal strategy would be to create an NLI dataset from scratch, with speakers of those languages creating the resource directly, ensuring that it represents culturally significant topics and inferences. However, this technique is significantly more expensive and time consuming than manually translating the data set and is, in most situations, impracticable. This is because finding fluent bilingual speakers to do 10,000 translations for all 11 languages is very hard. 

%% file: experiments_v2.tex
\section{Experiments}
Our experiments compare the performance of several multi-lingual models, including one particularly developed for \emph{Indic} languages. We consider 2 broad categories, \begin{inparaenum}[(a)] \item \textbf{\textit{Indic} Specific} which includes IndicBERT and MuRIL due to their indic specific pretraining, and \item \textbf{Generic} which includes mBERT and XLM-Roberta due to their pretraining in more than 100 languages. \end{inparaenum} We fine-tuned pre-trained multi-lingual models to develop NLI classifiers. The classifiers takes two sentence as input, i.e. the premise and the hypothesis and predicts the inference label. See appendix \S\ref{sec:hyperparameters_appendix} for models and hyper-parameters details respectively. 

\subsection{Experimental Setup}
In this section, we further elaborate upon categories of models used and training strategies employed.

\subsubsection{Details: Multi-lingual Models}
\label{sec: multilingual models appendix}

We explore two categories of multilingual models in our experiments, as detailed below: 

\paragraph{\emph{Indic} Specific:} These models are specially pre-trained using Mask Language Modeling (MLM) or Translation Language Model (TLM) \citep{conneau2019cross} on monolingual / bilingual \emph{Indic} language corpora.  These include models such as MuRIL and IndicBERT trained on 17 and 11 \emph{Indic} languages ($+$English) respectively. MuRIL is pre-trained using Common-Crawl Oscar Corpus \citep{OrtizSuarezSagotRomary2019}, PMIndia \citep{haddow2020pmindia} on the following languages: \emph{en, hi, bn, gu, te, ta, or, ml, pa, kn, mni, as, ur}. IndicBERT is pre-trained using \emph{Indic}-Corp\cite{kakwani-etal-2020-indicnlpsuite} on the following languages: \emph{en, hi, bn, ta, ml, te, Mr, kn, gu, pa, or, as}. Moreover, MuRIL is also pre-trained with TLM objective (with MLM objective) on machine translated data and machine transliterated data.

\paragraph{\emph{Generic}:} These are massive multi-lingual models pre-trained on large number of languages with MLM. These include multi-lingual BERT i.e. mBERT (cased/uncased) and multi-lingual RoBERTa i.e. XLM-RoBERTa which are trained on more than 100 languages. XLM-RoBERTa also includes pre-training on all eleven \emph{Indic} languages. XLM-RoBERTa is pre-trained using the common crawl monolingual data. mBERT (cased/uncased) includes pre-training on nine of eleven \emph{Indic} languages (Assamese and Odia excluded) and uses multi-lingual Wikipedia data for pre-training.

\subsubsection{Training-Evaluation Strategies.} 
To train the NLI classifier, we investigate several strategies. 
\begin{enumerate}  

\item \textbf{\emph{Indic} Train:} The models are trained and evaluated on  \datasetName. The training set is translated from the XNLI English, thus a \textit{translate-train} scenario. 

\item \textbf{English Train:} The models are trained on original English {\sc XNLI} data and evaluated on \datasetName data. This is a \textit{zero-shot evaluation} training scenario. 
\item \textbf{English Eval:} The model are trained on original English {\sc XNLI}  data, but evaluated on English translation of \datasetName data. This is the \textit{translate-test} scenario.

\item \textbf{English + \emph{Indic} Train:} This approach combines approaches \emph{(1)} and \emph{(2)}. The model is first pre-finetuned \citep{lee-etal-2021-meta,aghajanyan2021muppet} on English {\sc XNLI} data and then finetuned on \textbf{\emph{Indic} language} of \datasetName data. 

\item \textbf{Train All:} This approach begins by fine-tuning the pre-trained model on English {\sc XNLI} data, followed by training on \textit{all eleven \emph{Indic} languages} of \datasetName sequentially. 

\item \textbf{Cross Lingual Transfer:} Additionally, we assess the models' capacity to transfer between languages. Where the model is trained on a single Indian language and then assessed on all other Indian languages as well as the training language.

\item \textbf{Intra-Bilingual Inference:} Lastly, We also asses the model's capability to perform natural language inference with premise in English and hypothesis in Indic language. 
\end{enumerate} 

While the pre-trained multi-lingual models remain constant, the training and evaluation datasets vary.

\input{table_parta_v2}

%% file: table_parta_v2.tex
\begin{table*}[!htbp]
\centering
\footnotesize
\begin{tabular}{m{3.5em}|m{0.3em}m{0.3em}m{0.3em}m{0.3em}m{0.3em}m{0.3em}m{0.3em}m{0.3em}m{0.3em}m{0.3em}m{0.7em}|m{3.2em}<{\centering} | m{0.3em}m{0.3em}m{0.3em}m{0.3em}m{0.3em}m{0.3em}m{0.3em}m{0.3em}m{0.3em}m{0.3em}m{0.7em}|m{3.3em}<{\centering}}
\toprule \\[-1.2em]
\multirow{2}{*}{\textbf{Model}} & \multicolumn{11}{c|}{\textbf{Indic Train}} &  \multirow{2}{*}{\textbf{ModAvg}} & \multicolumn{11}{c|}{\textbf{English Train}} &  \multirow{2}{*}{\textbf{ModAvg}} \\ \\[-1.2em]
\cmidrule(lr){2-12} \cmidrule(lr){14-24} \\[-1.2em]
 & \textbf{as} & \textbf{gu} & \textbf{kn} &\textbf{ml} & \textbf{mr}& \textbf{or} & \textbf{pa} & \textbf{ta} & \textbf{te} & \textbf{bn} & \textbf{hi} & & \textbf{as} & \textbf{gu} & \textbf{kn} &\textbf{ml} & \textbf{mr}& \textbf{or} & \textbf{pa} & \textbf{ta} & \textbf{te} & \textbf{bn} & \textbf{hi} &  \\ \midrule \\ \\[-2.3em]
 \textbf{XLM-R} &\textcolor{BrickRed}{\textbf{70}} &73 &\textcolor{BrickRed}{\textbf{75}} &70 & \textcolor{BrickRed}{\textbf{75}} &32 &71 &76 &76 &\textcolor{BrickRed}{\textbf{76}} &\textcolor{ForestGreen}{\textbf{78}} &70 &65 & \textcolor{BrickRed}{\textbf{66}} &69 & \textcolor{BrickRed}{\textbf{69}} &67 & \textcolor{BrickRed}{\textbf{67}} &61 &71 &69 &69 &\textcolor{blue}{\textbf{73}} & 69 \\ \\[-1.2em]
 \textbf{iBERT} &67 &69 &68 &60 &68 &69 & \textcolor{ForestGreen}{\textbf{73}} &37 &62 &70 &68 &65 &57 &63 &53 &42 &59 &57 &\textcolor{blue}{\textbf{66}} &41 &56 &48 &63 &60 \\ \\[-1.2em]
\textbf{mBERT} &71 &62 &69 &71 &71 &35 &70 &70 &69 &67 & \textcolor{blue}{\textbf{74}} &66 &51 &57 &57 &57 &54 &34 &59 & \textcolor{blue}{\textbf{61}} &59 &57 &67 &59 \\ \\[-1.2em]
 \textbf{MuRIL} &70 & \textcolor{ForestGreen}{\textbf{78}} & \textcolor{BrickRed}{\textbf{75}} &\textcolor{BrickRed}{\textbf{76}} &70 &\textcolor{BrickRed}{\textbf{76}} &72 & \textcolor{BrickRed}{\textbf{74}} & \textcolor{ForestGreen}{\textbf{78}} &75 &71 &\textbf{74} &\textcolor{BrickRed}{\textbf{68}} &32 & \textcolor{BrickRed}{\textbf{75}} &34 & \textcolor{BrickRed}{\textbf{68}} &\textcolor{BrickRed}{\textbf{67}} &\textcolor{BrickRed}{\textbf{70}} &\textcolor{BrickRed}{\textbf{74}} &\textcolor{BrickRed}{\textbf{71}} &\textcolor{BrickRed}{\textbf{74}} &\textcolor{ForestGreen}{\textbf{76}} &\textbf{72} \\ \\[-1.2em]
\hdashline \\[-1.2em]
 \textbf{LangAvg} &68 &69 &70 &69 &70 &49 &71 &65 &70 &70 &\textbf{72} &68 &58 &55 &64 &52 &61 &52 &63 &61 &63 &62 &\textbf{68} &63 
\end{tabular}

\begin{tabular}{m{3.5em}|m{0.3em}m{0.3em}m{0.3em}m{0.3em}m{0.3em}m{0.3em}m{0.3em}m{0.3em}m{0.3em}m{0.3em}m{0.7em}|m{3.2em}<{\centering} | m{0.3em}m{0.3em}m{0.3em}m{0.3em}m{0.3em}m{0.3em}m{0.3em}m{0.3em}m{0.3em}m{0.3em}m{0.7em}|m{3.3em}<{\centering}}
\midrule \\[-1.2em]
\multirow{2}{*}{\textbf{Model}} & \multicolumn{11}{c|}{\textbf{English Eval}} & \multirow{2}{*}{\textbf{ModAvg}} & \multicolumn{11}{c|}{\textbf{English+\emph{Indic} Train}} &  \multirow{2}{*}{\textbf{ModAvg}}
 \\ \\[-1.2em]
\cmidrule(lr){2-12} \cmidrule(lr){14-24} \\[-1.2em]
& \textbf{as} & \textbf{gu} & \textbf{kn} &\textbf{ml} & \textbf{mr}& \textbf{or} & \textbf{pa} & \textbf{ta} & \textbf{te} & \textbf{bn} & \textbf{hi} & & \textbf{as} & \textbf{gu} & \textbf{kn} &\textbf{ml} & \textbf{mr}& \textbf{or} & \textbf{pa} & \textbf{ta} & \textbf{te} & \textbf{bn} & \textbf{hi} &  \\ \midrule \\ \\[-2.3em]

\textbf{XLM-R}  & \textcolor{BrickRed}{\textbf{66}} & \textcolor{BrickRed}{\textbf{72}} & 70 & \textcolor{BrickRed}{\textbf{68}} & \textcolor{BrickRed}{\textbf{66}} & 65 & \textcolor{BrickRed}{\textbf{72}} & 69 & \textcolor{BrickRed}{\textbf{72}} & 71 &\textcolor{blue}{\textbf{75}}&\textbf{70} & \textcolor{BrickRed}{\textbf{73}} & 75 & \textcolor{BrickRed}{\textbf{77}} & 75 & \textcolor{BrickRed}{\textbf{74}} & 73 & 75 & 75 & 73 & \textcolor{BrickRed}{\textbf{75}} & \textcolor{ForestGreen}{\textbf{79}} & 76 \\ \\[-1.2em]
\textbf{iBERT} & 63 & 66 & 68 & 61 & 65 & 65 & 66 & 63 & 63 &\textcolor{blue}{\textbf{72}} &\textcolor{blue}{\textbf{72}}& 66 & 67 & 72 & 65 & 62 & 59 & 59 &\textcolor{blue}{\textbf{74}} & 63 & 66 & 69 &\textcolor{blue}{\textbf{74}}&70 \\ \\[-1.2em]
\textbf{mBERT} & 62 & 64 & 67 & 65 & 61 & 60 & 66 & \textcolor{BrickRed}{\textbf{67}} & 66 &\textcolor{blue}{\textbf{75}} & 72 & 66 & 67 & 70 & 69 & 70 & 70 & 39 & 71 &\textcolor{blue}{\textbf{73}} & 70 & 70 &71 & 69 \\ \\[-1.2em]
\textbf{MuRIL} & 65 & 33 & \textcolor{BrickRed}{\textbf{71}} & 67 & 67 & \textcolor{BrickRed}{\textbf{67}} & 71 & 31 & 71 & 72 & \textcolor{ForestGreen}{\textbf{77}} & 63 & 76 & \textcolor{BrickRed}{\textbf{77}} & 77 & \textcolor{ForestGreen}{\textbf{79}} & 74 & \textcolor{BrickRed}{\textbf{76}} &\textcolor{ForestGreen}{\textbf{77}} &\textcolor{ForestGreen}{\textbf{77}}& \textcolor{BrickRed}{\textbf{74}} & \textcolor{BrickRed}{\textbf{75}} &\textcolor{blue}{\textbf{77}}& \textbf{77} \\ \\[-1.2em]
\hdashline \\[-1.2em]
\textbf{LangAvg} & 64 & 60 & 68 & 65 & 63 & 64 & 69 & 60 & 68 & 73 &\textbf{74}& 66 & 69 & 73 & 70 & 72 & 68 & 56 & 73 & 72 & 70 & 72 &\textbf{75} & \textbf{72} 
\end{tabular}

\begin{tabular}{m{3.5em}|m{0.3em}m{0.3em}m{0.3em}m{0.3em}m{0.3em}m{0.3em}m{0.3em}m{0.3em}m{0.3em}m{0.3em}m{0.7em}|m{3.2em}<{\centering} | m{0.3em}m{0.3em}m{0.3em}m{0.3em}m{0.3em}m{0.3em}m{0.3em}m{0.3em}m{0.3em}m{0.3em}m{0.7em}|m{3.3em}<{\centering}} 
\midrule \\[-1.2em]
\multirow{2}{*}{\textbf{Model}} & \multicolumn{11}{c|}{\textbf{Train All}} & \multirow{2}{*}{\textbf{ModAvg}} & \multicolumn{11}{c|}{\textbf{Cross Lingual Transfer}} & \multirow{2}{*}{\textbf{ModAvg}} \\ \\[-1.2em]
\cmidrule(lr){2-12} \cmidrule(lr){14-24} \\[-1.2em]
& \textbf{as} & \textbf{gu} & \textbf{kn} &\textbf{ml} & \textbf{mr}& \textbf{or} & \textbf{pa} & \textbf{ta} & \textbf{te} & \textbf{bn} & \textbf{hi} & & \textbf{as} & \textbf{gu} & \textbf{kn} &\textbf{ml} & \textbf{mr}& \textbf{or} & \textbf{pa} & \textbf{ta} & \textbf{te} & \textbf{bn} & \textbf{hi} &  \\ \midrule \\ \\[-2.3em]
\textbf{XLM-R}  &\textcolor{BrickRed}{\textbf{73}} &\textcolor{ForestGreen}{\textbf{77}} &\textcolor{BrickRed}{\textbf{74}} &\textcolor{BrickRed}{\textbf{76}} &72 &73 &\textcolor{blue}{\textbf{77}} &\textcolor{blue}{\textbf{77}}& \textcolor{BrickRed}{\textbf{76}} &\textcolor{blue}{\textbf{77}}&\textcolor{blue}{\textbf{77}} &75 & 66 & 70 & 33 & 34 & 70 & 35 & 68 & 70 & 70 & 71 & \textcolor{blue}{\textbf{72}} & 60 \\ \\[-1.2em]
\textbf{iBERT} & 63 & 74 & 59 & 51 & 69 & 66 &\textcolor{blue}{\textbf{75}} & 60 & 67 & 70 & 74 & 66 & 59 & \textcolor{blue}{\textbf{60}} & 59 & 54 & \textcolor{blue}{\textbf{60}} & \textcolor{ForestGreen}{\textbf{60}} & \textcolor{blue}{\textbf{60}} & 56 & 59 & 58 & \textcolor{blue}{\textbf{60}} & 59 \\ \\[-1.2em]
\textbf{mBERT} & 63 & 69 & 69 & 71 & 70 & 33 & 71 & 69 & 70 &\textcolor{blue}{\textbf{74}} & 72 & 66 & 57 & 59 & \textcolor{blue}{\textbf{60}} & 59 & 58 & 33 & 59 &\textcolor{blue}{\textbf{60}} & 59 & \textcolor{blue}{\textbf{60}} & \textcolor{blue}{\textbf{60}} & 57 \\ \\[-1.2em]
\textbf{MuRIL} & \textcolor{BrickRed}{\textbf{73}} & 76 & \textcolor{BrickRed}{\textbf{74}} & \textcolor{BrickRed}{\textbf{76}} & \textcolor{BrickRed}{\textbf{74}} & \textcolor{BrickRed}{\textbf{78}} &\textcolor{ForestGreen}{\textbf{81}} & \textcolor{BrickRed}{\textbf{78}} & \textcolor{BrickRed}{\textbf{76}} & \textcolor{BrickRed}{\textbf{80}} & \textcolor{BrickRed}{\textbf{78}} &\textbf{77} & \textcolor{BrickRed}{\textbf{75}} & \textcolor{BrickRed}{\textbf{73}} & \textcolor{BrickRed}{\textbf{75}} &\textcolor{ForestGreen}{\textbf{76}}& \textcolor{BrickRed}{\textbf{71}} & 33 & \textcolor{BrickRed}{\textbf{75}} & \textcolor{ForestGreen}{\textbf{76}} & \textcolor{BrickRed}{\textbf{73}} & \textcolor{BrickRed}{\textbf{75}} & \textcolor{BrickRed}{\textbf{73}} & \textbf{70} \\ \\[-1.2em]
\hdashline \\[-1.2em]
\textbf{LangAvg} & 68 & 73 & 69 & 68 & 71 & 58 & 75 & 71 & 71 &\textbf{75} & 74 & 70 & 63 & 64 & 57 & 56 & 63 & 39 & 64 & 64 & 64 &\textbf{65} &\textbf{65} & 60 \\
\bottomrule
\end{tabular}
\caption{\footnotesize Here, LangAvg represents the language wise average score across models, while ModAvg average score represents the model average score across languages. Values in \textcolor{blue}{\textbf{Blue}}, \textcolor{BrickRed}{\textbf{Red}} and \textcolor{ForestGreen}{\textbf{Green}} represents the model average best score, language-wise average best score, and values where both model-wise and language-wise best score coincide. For Indic Cross Lingual Transfer, each row represent the average evaluation score of all \emph{Indic} language when trained on the column language. For more detailed cross-lingual transfer results refer to Appendix \S\ref{sec:indic_crosslingual_appendix}. iBERT stand for \emph{indic}BERT and XLM-R stand for XLM-RoBERTa.
}
\label{tab:RQ_1234}
\end{table*}

%% file: results_analysis.tex
\subsection{Results and Analysis.} 
We summarizes our findings from Table \S\ref{tab:RQ_1234} results across four categories: 

\paragraph{\bf Across Models:} In all experiments, MuRIL performs the best across all \emph{indic} languages except in English Eval setup. This can be attributed to \begin{inparaenum}[(a.)] \item  The large model size \item  indic-specific pre-training data, \item A Mixture of Masked Language Modeling (MLM), Translation Language Modeling (TLM), and \item use of transliterated data in pre-training. \end{inparaenum} XLM-RoBERTa beats MuRIL in rare scenarios, notably in which the model solely deals with English data (e.g. English Eval). XLM-RoBERTa outperforms MuRIL in such cases because it is better at assessing English than MuRIL, which is designed mostly for indic language. Additionally, we discover that, compared to XLM-RoBERTa, MuRIL indic-specific training further enhances the model's performance. Despite indic-specific pretraining, IndicBERT performs worse than mBERT. This can be attributed to the smaller size of the IndicBERT model, i.e. only 33M compared to 167M mBERT (c.f. Table \S\ref{tab: hyper-params} in appendix).  

\paragraph{\bf Across Language:} We see a strong positive correlation between language performance with their resource availability. Hindi and Bengali outperform, whereas Odia mostly underperform on majority of benchmarks. Low-resource languages such as Marathi, Assamese, and Kannada surprising also perform well. This can be attributed to the similarity of Marathi with Hindi script, Assamese with Bengali script, and Kannada with Tamil and Telugu scripts. This is discussed in detail in appendix \ref{sec: error_analysis}. Odia, a low resource language, lacks script sharing language partners and hence performs poorly. Overall, English + \textit{Indic} Train method outperforms, with MuRIL performing best.

\paragraph{\bf Across Strategies:} Our experiments show that models benefit from language-specific fine tuning. English + \textit{Indic} train and Train All have the best results with minimal deviation across languages for XLM-R and MuRIL. Additionally, \emph{Train All} follows a high-to-low resource hierarchy to mitigate the impact of catastrophic forgetting \citep{goodfellow2015empirical}. Due to the followed language order English + \textit{Indic} train outperform Train All setting marginally for high resource languages. Overall, English + \textit{Indic} Train strategy performs the best and MuRIL performs the best in that strategy. This can be attributed to the \emph{indic} specific pre-training process of MuRIL which include both translation and transliteration. Furthermore, MuRIL has the second largest size after XLM-R. 

\paragraph{\bf Cross-Lingual Transfer:}\label{sec:indic_crosslingual} Models favour high resource languages such as \emph{Hindi} and \emph{Bengali} training for cross-lingual transfer. These language are pre-trained on large mono-lingual corpora which enhanced performance ~\citep{conneau2020unsupervised}. This setting can be thought equivalent of \emph{Hindi} and \emph{Bengali} substitution for English training. Additionally, when evaluated for all \emph{indic} languages, model trains on non-\emph{Hindi} and non-\emph{Bengali} perform substantially better for \emph{Hindi} and \emph{Bengali}. Table \ref{tab:RQ_1234} present results summary as average evaluation score across all \emph{indic} language(rows) when train on the several \emph{indic} languages (columns). \footnote{For model-wise cross-lingual results c.f. Appendix \S\ref{sec:indic_crosslingual_appendix}.} 

\paragraph{Intra-Bilingual Inference:}
\label{sec:english-Indic}
\input{tables_partc}

We also evaluate models on mixed input inference task {\sc{En-IndicXNLI}}, which consists of English \emph{premises}  paired with corresponding \emph{indic} hypothesis. We train model on mixed input using \textbf{English + \emph{Indic} Train} and \textbf{Train All} strategies. Table \ref{tab:RQ_6} shows performance of \textbf{English + \emph{Indic} Train} and \textbf{Train All} models on {\sc{En-IndicXNLI}}. Compared to uni-language inference task, mixed-language input task perform poorly. Furthermore, contrary to earlier observations, generic model such as XLM-R outperforms the \emph{Indic} specific models. 
However, IndicBERT and MuRIL both perform substantially better than mBERT. Furthermore, English data augmentation enhance the \textbf{English + \emph{Indic} Train} setting performance. This can be because, the model "meta-learns" the task successfully with English data training (premise language), and further prioritises the model's language-specific abilities with the follow-up indic data training.\\

\noindent \textit{Results Analysis.} We observed a performance loss except for XLM-RoBERTa when the model is evaluated on {\sc{En-IndicXNLI}} inference task. The inference models struggle to correlate and reason together on two different languages (English, \emph{Indic}) sentences. Contrary to earlier observation, a generic model such as XLM-RoBERTa outperforms the \emph{Indic} specific models. However, IndicBERT and MuRIL perform better than mBERT. \emph{Bengali} perform best for both the training strategies. We also observe the benefit of English data augmentation \textbf{English + \emph{Indic} Train} model, rather than all language augmentation \textbf{Train All} model.

%% file: tables_partc.tex
\begin{table*}[!htbp]
\centering
\small
\begin{tabular}{m{3.5em}|m{0.3em}m{0.3em}m{0.3em}m{0.3em}m{0.3em}m{0.3em}m{0.3em}m{0.3em}m{0.3em}m{0.3em}m{0.7em}|m{3.2em}<{\centering} | m{0.3em}m{0.3em}m{0.3em}m{0.3em}m{0.3em}m{0.3em}m{0.3em}m{0.3em}m{0.3em}m{0.3em}m{0.7em}|m{3.3em}<{\centering}}
\toprule \\[-1.2em]
\multirow{2}{*}{\textbf{Model}} & \multicolumn{11}{c|}{\textbf{English+\textit{Indic} Train}} &  \multirow{2}{*}{\textbf{ModAvg}} & \multicolumn{11}{c|}{\textbf{Train All}} &  \multirow{2}{*}{\textbf{ModAvg}} \\ \\[-1.2em]
\cmidrule(lr){2-12} \cmidrule(lr){14-24} \\[-1.2em]
& \textbf{as} & \textbf{gu} & \textbf{kn} &\textbf{ml} & \textbf{mr}& \textbf{or} & \textbf{pa} & \textbf{ta} & \textbf{te} & \textbf{bn} & \textbf{hi} & & \textbf{as} & \textbf{gu} & \textbf{kn} &\textbf{ml} & \textbf{mr}& \textbf{or} & \textbf{pa} & \textbf{ta} & \textbf{te} & \textbf{bn} & \textbf{hi} &\\\midrule
\textbf{XLM-R}  & \textcolor{BrickRed}{\textbf{74}} & \textcolor{BrickRed}{\textbf{72}} & \textcolor{BrickRed}{\textbf{75}} & \textcolor{BrickRed}{\textbf{74}} & \textcolor{ForestGreen}{\textbf{77}} & \textcolor{BrickRed}{\textbf{72}} & 70 & \textcolor{BrickRed}{\textbf{72}} & \textcolor{BrickRed}{\textbf{72}} & \textcolor{ForestGreen}{\textbf{79}} & \textcolor{BrickRed}{\textbf{76}} &\textbf{74} & \textcolor{BrickRed}{\textbf{57}} & \textcolor{BrickRed}{\textbf{59}} & \textcolor{BrickRed}{\textbf{58}} & \textcolor{BrickRed}{\textbf{62}} & \textcolor{BrickRed}{\textbf{61}} & 53 & 57 & 59 & 
\textcolor{BrickRed}{\textbf{61}} &\textcolor{ForestGreen}{\textbf{63}} &\textcolor{ForestGreen}{\textbf{63}} &\textbf{59} \\
\textbf{iBERT} & 70 & 68 & 63 & 65 & 69 & 68 &\textcolor{blue}{\textbf{71}} & 64 & 64 & 69 & 69 & 67 & 49 & 53 & 46 & 37 & 52 & 51 & 59 & 39 & 51 &\textcolor{blue}{\textbf{57}} & 50 & 50 \\
\textbf{mBERT} & 51 & 56 & 59 & 50 & 62 & 31 & \textcolor{blue}{\textbf{63}} & 57 & 60 & 61 &\textcolor{blue}{\textbf{63}} & 56 & 39 & 39 & 43 & 38 & \textcolor{blue}{\textbf{43}} & 33 & 40 & 42 & 41 & 40 & 42 & 40 \\
\textbf{MuRIL} & 71 & 70 &\textcolor{blue}{\textbf{73}} & 69 & 71 & 39 & 71 & 71 & 69 & 72 & 69 & 67 & 51 & 52 & 58 & 56 & 53 & \textcolor{BrickRed}{\textbf{55}} & \textcolor{BrickRed}{\textbf{58}} &\textcolor{blue}{\textbf{65}} & 55 & 62 & 54 & 56 \\
\hdashline
\textbf{LangAvg} & 65 & 65 & 66 & 64 & 68 & 53 & 62 & 65 & 67 &\textbf{71} & 70 & 65 & 47 & 49 & 51 & 48 & 51 & 45 & 52 & 50 & 51 &\textbf{54} & 51 &50\\
\midrule
\bottomrule
\end{tabular}
\caption{{{\sc En-IndicXNLI}} model performance (refer \S \ref{sec:english-Indic}) with {English + \emph{Indic} train} and {Train All} setting. Here, ModAvg, LangAvg, and Color Code mean same as in {table \ref{tab:RQ_1234}}.}
\label{tab:RQ_6}
\end{table*}


%% file: error_analysis.tex
\subsection{Error Analysis}
\label{sec: error_analysis}
\input{error_analysis_figures}
In this section we investigate the correlation between the language similarity and the model performance. We see that the model performs similarly on similar languages. We evaluate our results on MuRIL on the  English+\emph{Indic} finetuning Strategy.

\noindent In Figure \ref{fig:consistency_matrix}, we observe that the overall Correct and Incorrect predictions, Bengali vs Assamese pair has the total of 81\% overlap,Tamil vs Kannada has 83\% overlap, Hindi vs Maratha has 82\% overlap. All the language pairs have the largest overlap for entailment label for correct labels and largest overlap in contradiction label for incorrect overlaps.

\noindent In Figure \ref{fig:confusion_matrix}, interestingly Bengali vs Assamese pair and Hindi vs Marathi has the highest percentage of overlap in predictions where the most overlap is in entailment and minimum overlap is in contradiction. While for Tamil vs Kannada pair has the highest overlap for neutral and minimum for contradiction.\\

We have also done error analysis of model performance on original hindi test data already present in XNLI and data obtained through translations from IndicTrans in Figure \ref{fig:consistency_hi_hi_orig}. We observe that there is a total of 82\% of overlap in error consistency and we again observe that the most number of correct overlaps in for entailment label and most number of incorrect predictions are for contradiction label. In terms of consistency, we see the maximum overlap in neutral prediction and least overlap in contradiction prediction. This shows that model performs similarly on the original Hindi data and machine translated Hindi data enhancing the validity of the \datasetName dataset.

%% file: error_analysis_figures.tex
\begin{figure*}
    \vspace{-2.0em}
    \subfloat[Tamil vs Kannada]{\includegraphics[scale=0.3]{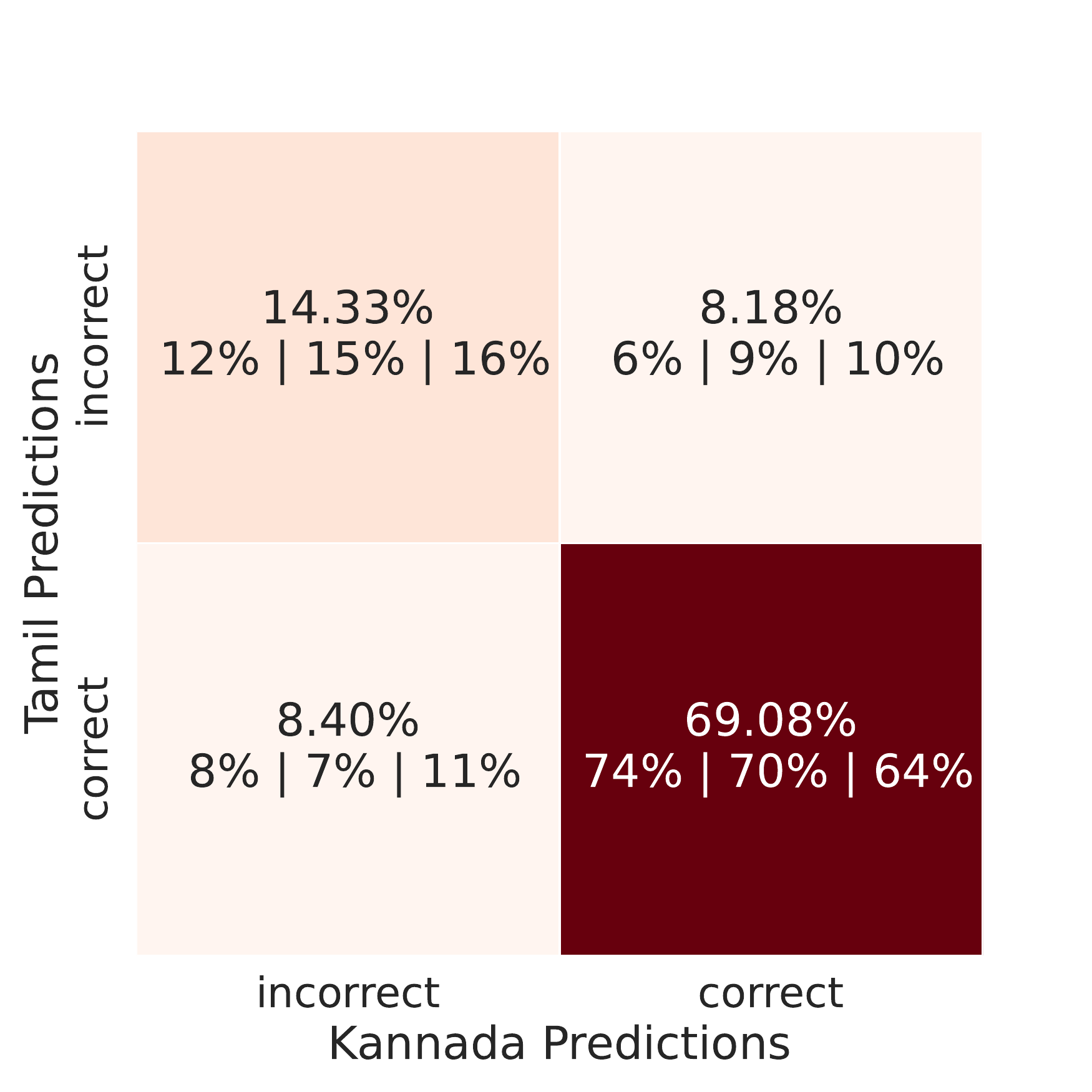}}
    \subfloat[Bengali vs Assamese]{\includegraphics[scale=0.3]{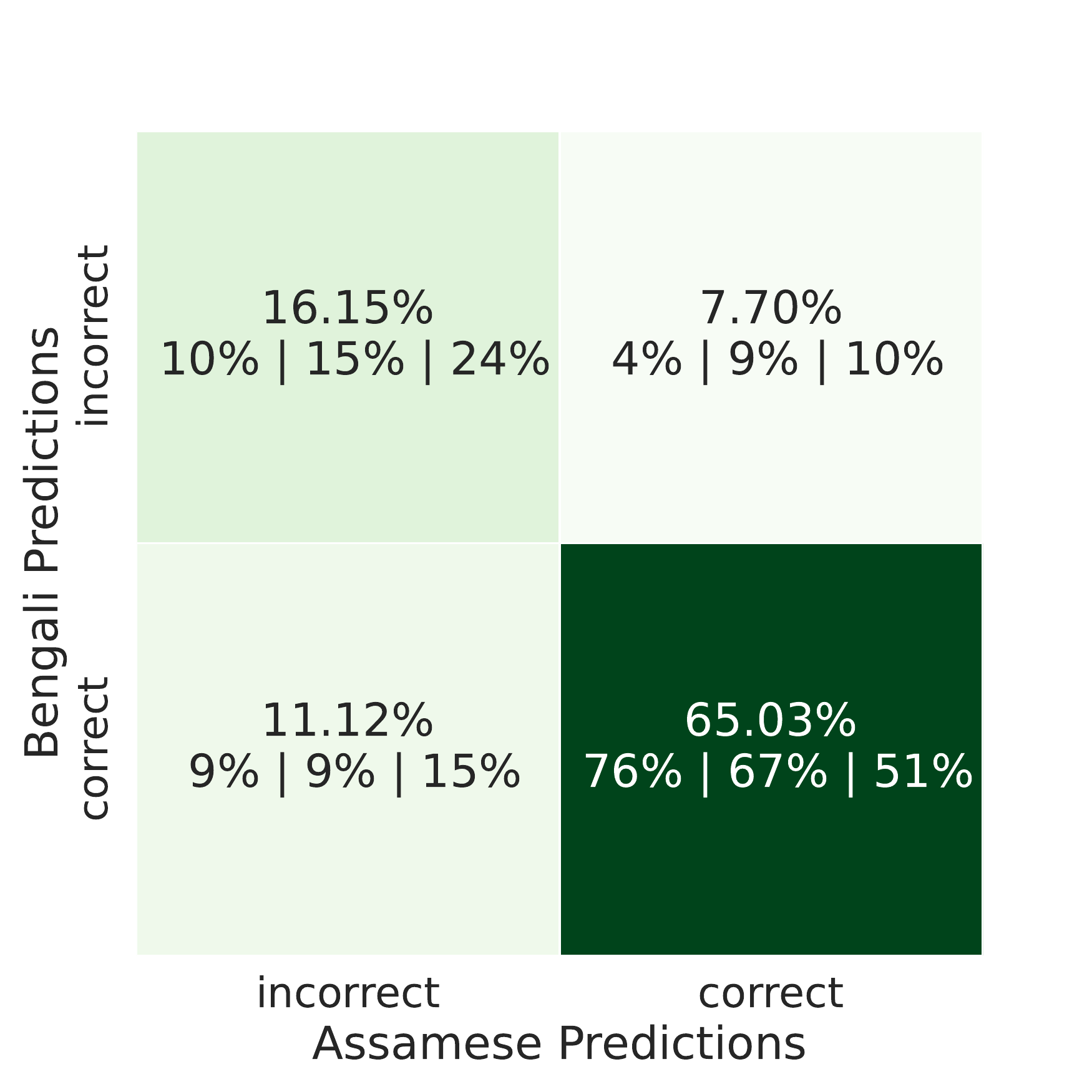}}
    \subfloat[Hindi vs Marathi]{\includegraphics[scale=0.3]{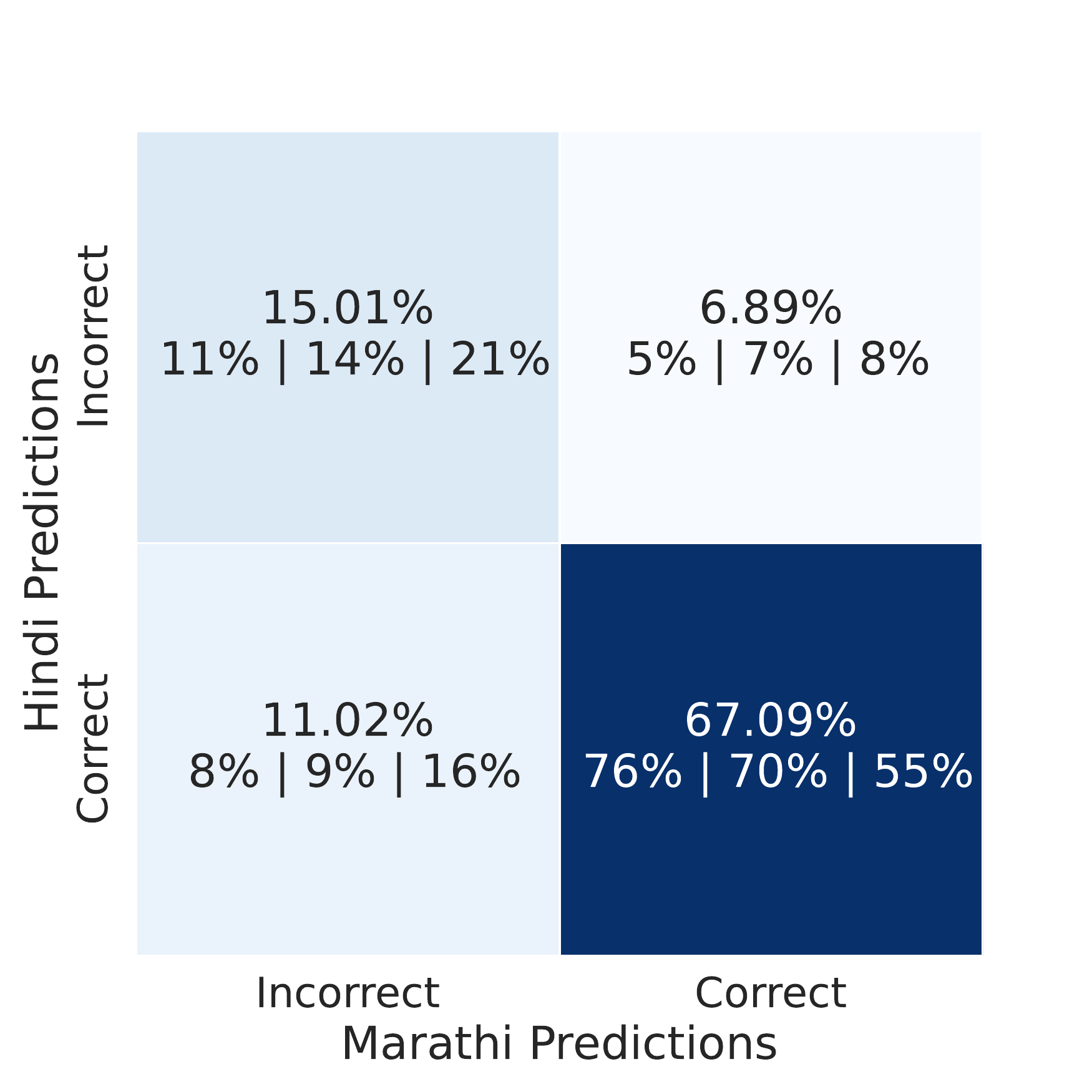}}
    \caption{Consistency Matrix: Predictions of MuRIL for (a) Tamil vs Kannada (b) Bengali vs Assamese, , (c) Hindi vs Marathi. The percentage on top in each block represents the average across all three labels with each label percentage given below it in the order of Entailment, Neutral and Contradiction.}
    \label{fig:consistency_matrix}
\end{figure*}

\begin{figure*}
    \vspace{-2.0em}
    \subfloat[Tamil vs Kannada]{\includegraphics[scale=0.3]{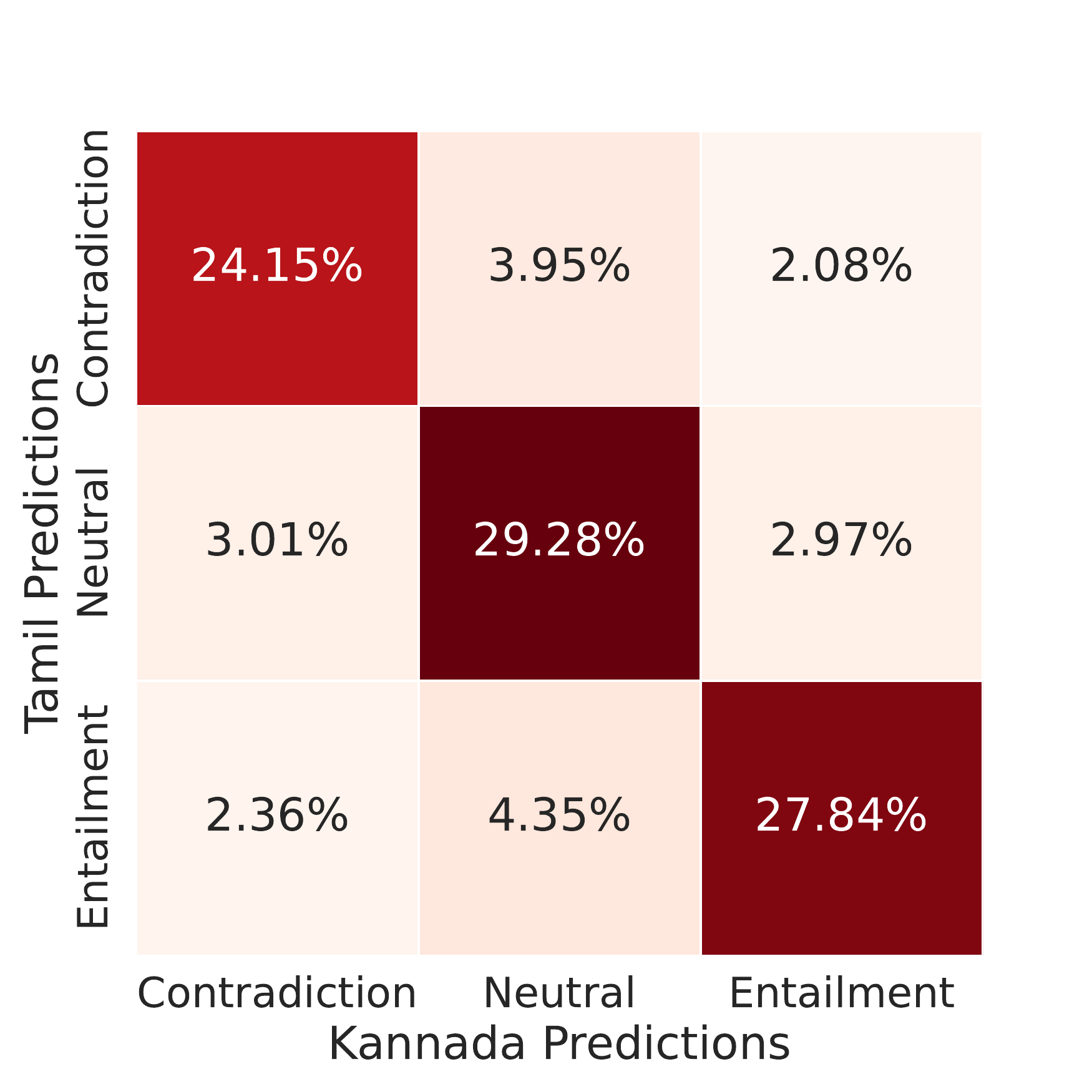}}
    \subfloat[Bengali vs Assamese]{\includegraphics[scale=0.3]{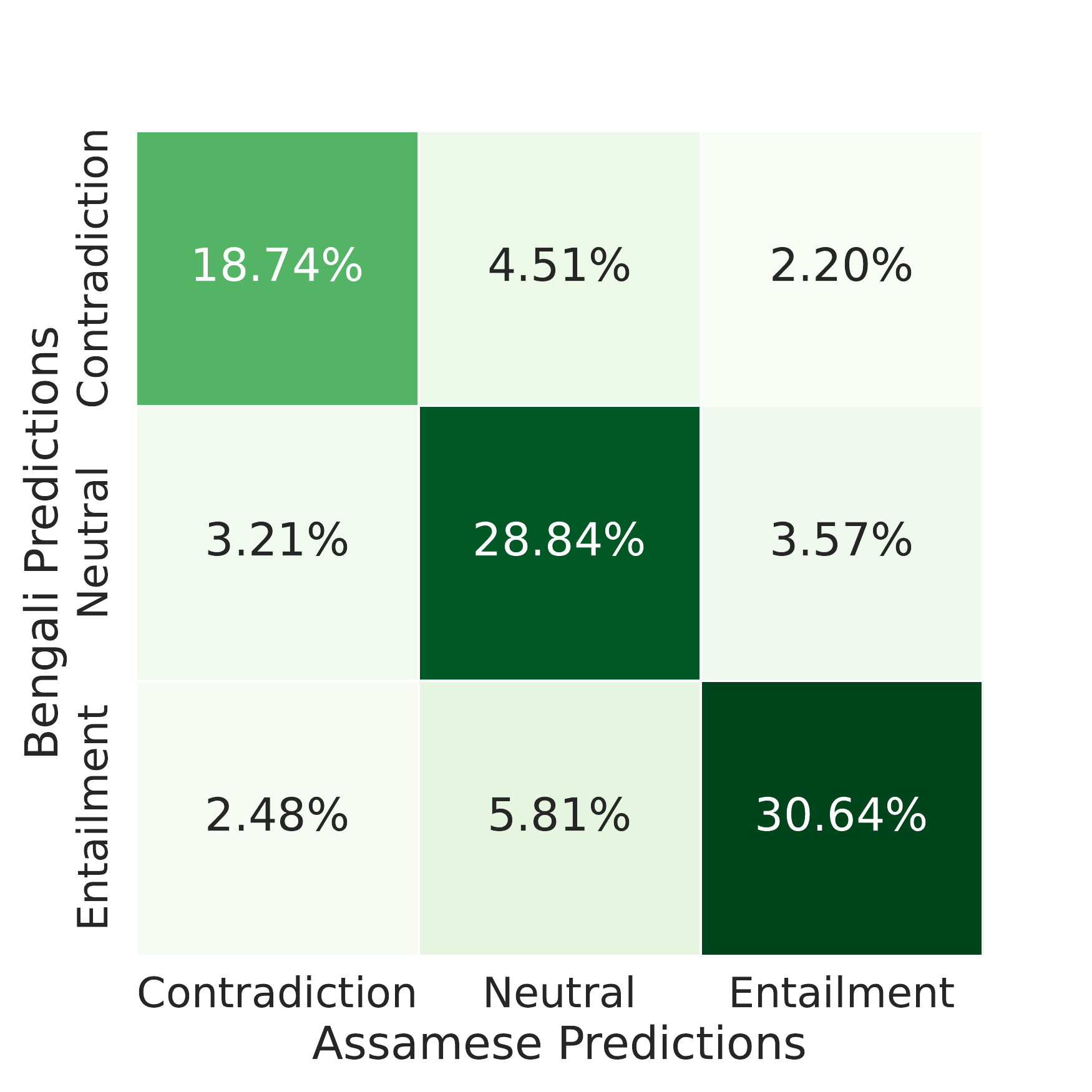}}
    \subfloat[Hindi vs Marathi]{\includegraphics[scale=0.3]{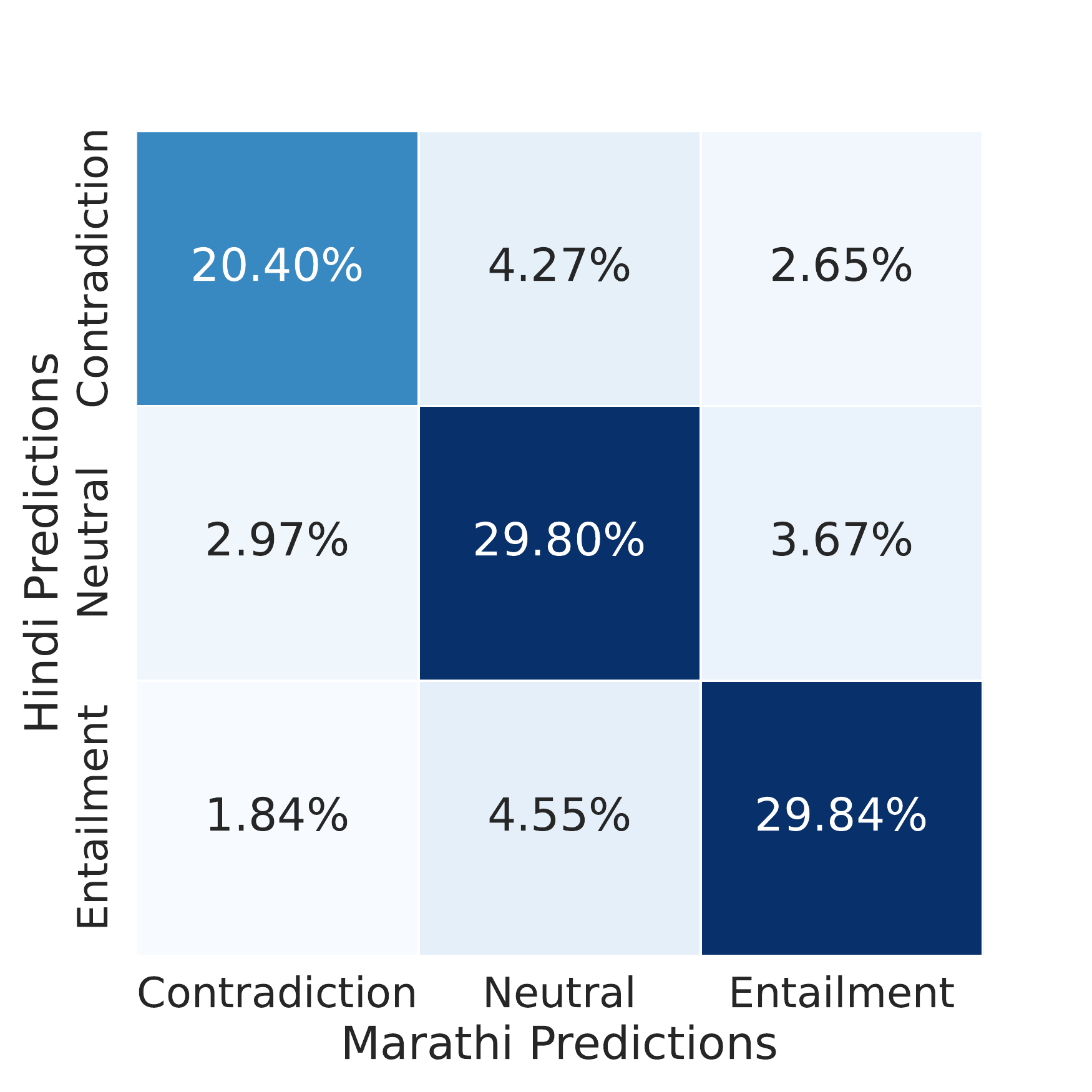}}
    \caption{Confusion Matrix: for MuRIL (a) Tamil vs Kannada, (b) Bengali vs Assamese,  (c) Hindi vs Marathi.}
    \label{fig:confusion_matrix}
\end{figure*}

\begin{figure*}
    \vspace{-2.0em}
    \centering
    \subfloat[Confusion Matrix]{\includegraphics[scale=0.35]{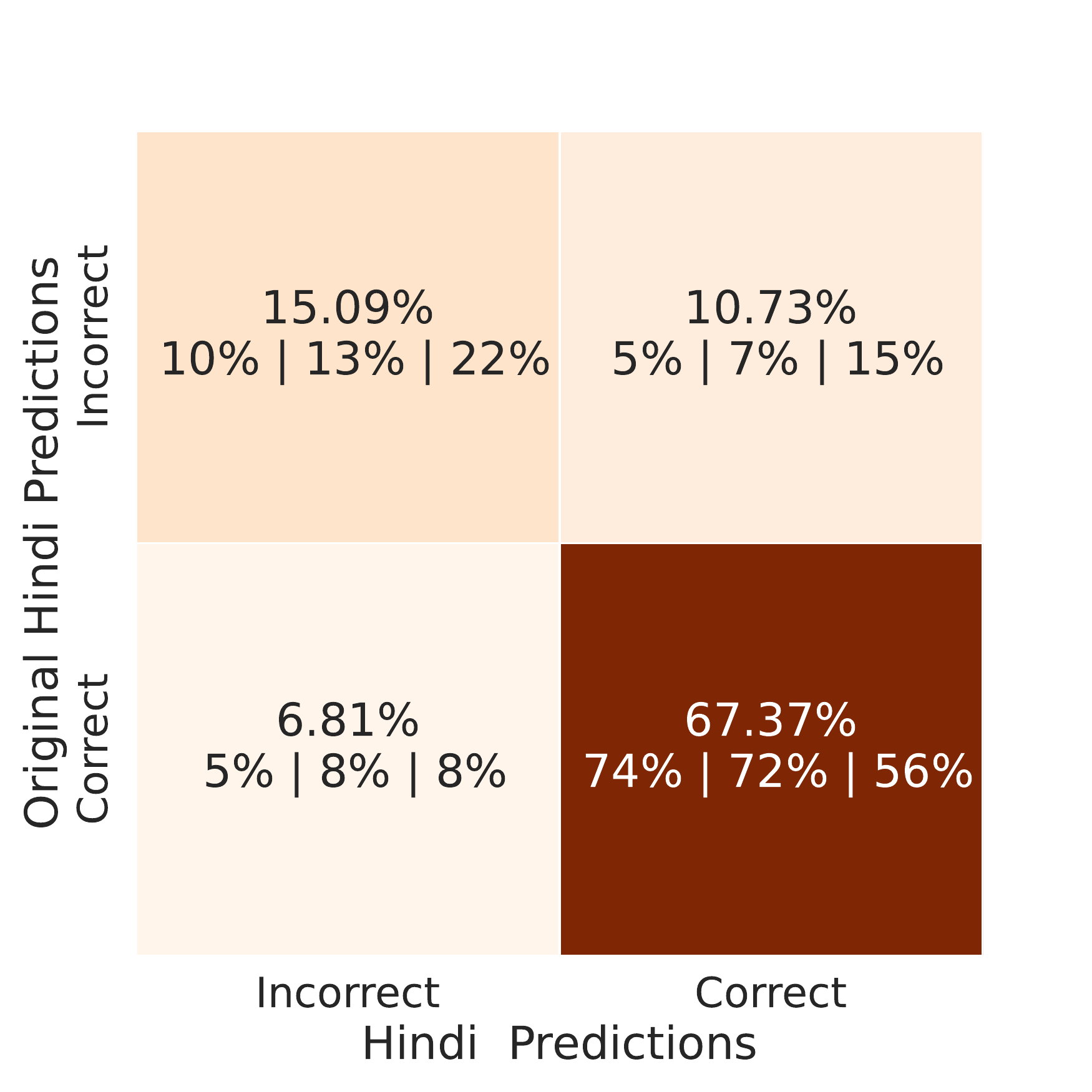}}
    \subfloat[Consistency Matrix]{\includegraphics[scale=0.35]{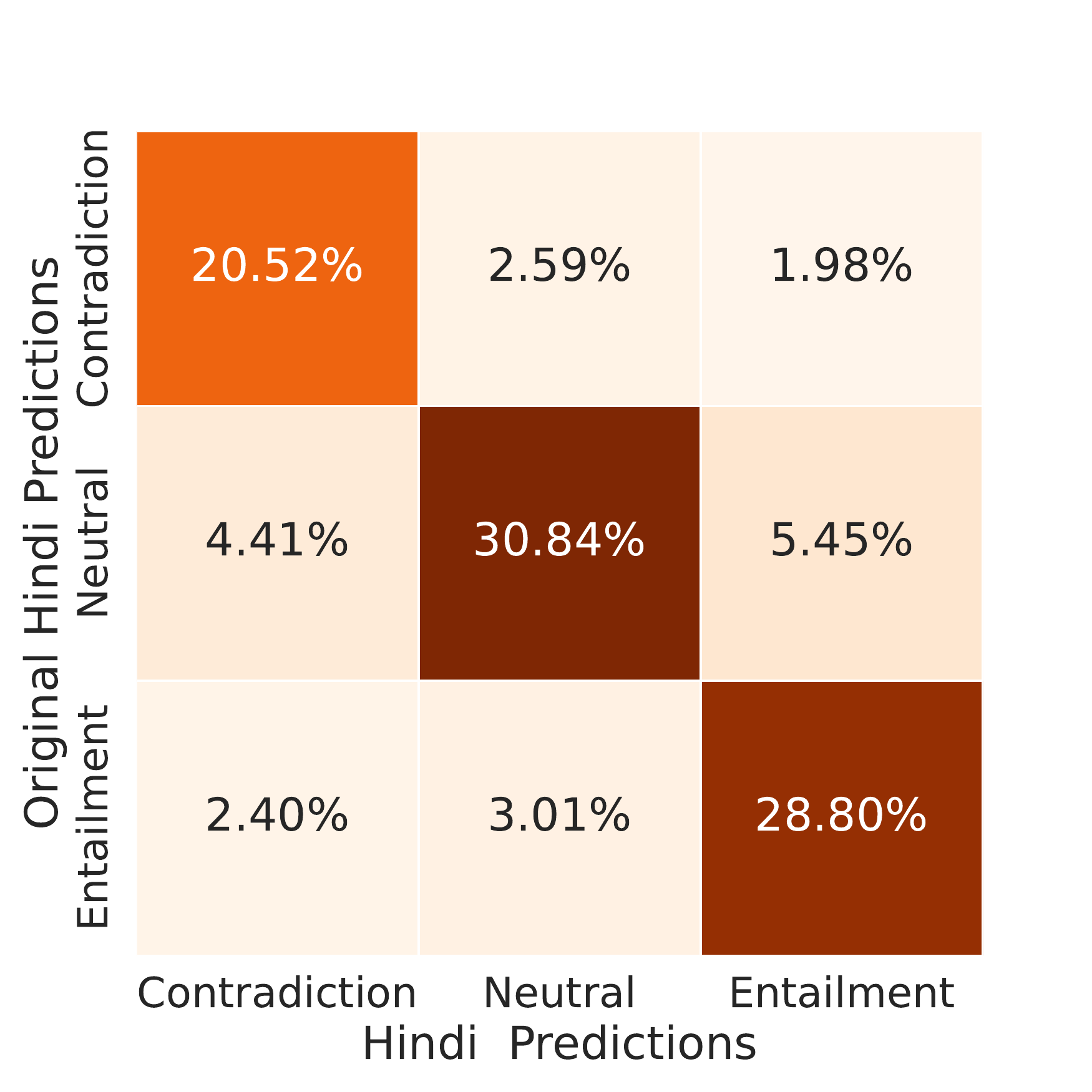}}
    \caption{Consistency Matrix and Confusion Matrix for Predictions of MuRIL on Original Hindi data in XNLI and Machine Translated Data generated from IndicTrans.}
    \label{fig:consistency_hi_hi_orig}
\end{figure*}

%% file: motivation.tex
\section{Discussion}
\label{sec:further-discussion}

\paragraph{Why \emph{Indic} languages?} Indic languages are spoken by more than a billion people in the Indian subcontinent. With the introduction of IndicNLPSuite \citep{kakwani-etal-2020-indicnlpsuite} by AI4Bharat\footnote{\url{https://ai4bharat.org}} there has been has an increased interest and effort towards the research for \emph{Indic} languages model. Recently, IndicBERT, MuRIL \citep{khanuja2021muril} based on BERT \citep{devlin2019bert} were introduced for the \emph{Indic} languages. Furthermore, generation model IndicTrans \citep{ramesh2021samanantar} and IndicBART \citep{dabre2021indicbart} based on seq2seq architecture was also published recently. These model use the \emph{Indic} enrich monolingual corpora: Common Crawl, Oscar and IndicCorp and parallel corpora: Samantar and PMIndia \citep{haddow2020pmindia} on \emph{Indic} languages for training.  Despite significant progress through large transformer-based \emph{Indic} language models in addition to existing multilingual models e.g. mBERT \citep{devlin2019bert} , XLM-RoBERTa \citep{conneau2020unsupervised}, and mBART (seq2seq) \cite{liu-etal-2020-multilingual-denoising} there is currently a paucity of benchmark data-sets for evaluating these huge language models in the \emph{Indic} language research field. Such benchmark dataset is necessary for studying the linguistic features of Indic languages and how well they are perceived by different multilingual models.  Recently, IndicGLUE \cite{kakwani-etal-2020-indicnlpsuite} was introduced to handle this scarcity. However, the scope of this benchmark, is confined to only few tasks and datasets.

\paragraph{Why \datasetName task?} This research provides an excellent chance to investigate the efficacy of various Multilingual models on \emph{Indic} languages that are rarely evaluated or explored before. Some of these \emph{Indic} languages such as \emph{`Assamese'} and \emph{`Odia'} serve as unseen (zero-shot) evaluation for models such as mBERT \citep{pires-etal-2019-multilingual}, i.e. not pre-trained on \emph{`Assamese'}. While other models, such as XLM-RoBERTa, IndicBERT and MuRIL covers all our languages but in widely varying proportions in their training data. Our work investigate the correlation effect of cross-lingual training for English on these rare \emph{Indic} languages, which are not explore by prior studies. \\

Furthermore, we also investigate the cross-lingual transfer effect across \emph{Indic} languages, also not explored before. We explore the impact of Multilingual training, english-data augmentation, unified Indic model performance, cross-lingual transfer of closely related \emph{Indic} family  and English-\emph{Indic} NLI through our work. All the above mention topics are not explore for \emph{Indic} language before. We aim to integrate \datasetName and benchmark models in IndicGLUE \cite{kakwani-etal-2020-indicnlpsuite}. Such a benchmark dataset is required for investigating the linguistic properties of Indian languages and how accurately they are interpreted by various multilingual models. Another direction is accessing model performance on {\sc Indic-IndicXNLI} task,  where both premises and hypothesis are in two distinct \emph{Indic} languages. 

\paragraph{Why IndicTrans for Translation?}
\label{why_indicTrans_appendix}

We use the IndicTrans as a translation model for converting English {\sc XNLI} to \datasetName because of the following reasons:

\begin{itemize}
\item \textbf{Open-Source:} IndicTrans is open-source to public for non-commercial usage without additional fees, while Google-Translate and Microsoft-Translate require paid subscription. 

\item \textbf{Light Weight:} IndicTrans is the fastest and the lightest amongst mBART and mT5 on single-core GPU machines. Google-Translate and Microsoft-Translate are also relatively slower due to repeated network-intensive API calls.

\item \textbf{\emph{indic} Coverage:} Seq2Seq models like mBART and mT5 are not designed for all languages in the indic family. mBART supports eight (excludes kn,or,pa,as) while mT5 supports nine languages (excludes or,as) out of eleven indic languages. Google-Translate supports ten out of eleven \emph{indic} languages (excludes Assamese). Microsoft Translate supports all the eleven \emph{indic} languages.
\end{itemize} 

In future, we plan to enhance \datasetName with better translation methods.

%% file: relatedwork.tex
\vspace{-0.3em}
\section{Related Work}
\label{sec:relatedework}
Recently many \emph{Indic}-specific resources are developed such as IndicNLPSuite \cite{kakwani-etal-2020-indicnlpsuite}, which include \begin{inparaenum}[(a.)] \item word embeddings: IndicFT, \item transformer models: IndicBERT, \item monolingual corpora: IndicCorp, \item and, evaluation benchmark: IndicGLUE \end{inparaenum}. Furthermore, \emph{Indic}-specific pre-processing libraries such as iNLTK \citep{arora-2020-inltk} and Indic-nlp-library \citep{kunchukuttan2020indicnlp}, other Indic monolingual corpora: Common Crawl Oscar Corpus \cite{wenzek-etal-2020-ccnet,ortiz-suarez-etal-2020-monolingual},  multilingual parallel corpora: PMIndia \cite{haddow2020pmindia} and Samantar \cite{ramesh2021samanantar}, transformer model MuRIL \cite{khanuja2021muril} and language specific Indic-Transformers \citep{jain2020indictransformers} exists.

\section{Conclusion}
\label{sec:conclusion}

With \datasetName we extend the {\sc XNLI} dataset for \emph{Indic} languages family. We benchmark \datasetName with several multi-lingual models using various train-test strategies. We also study the use of English XNLI as pre-finetuning dataset. Furthermore, we also evaluate models on mixed-language inference input and cross-lingual transfer ability. We aim to integrate \datasetName and benchmark models in IndicGLUE \cite{kakwani-etal-2020-indicnlpsuite}. We also intend to enhance \datasetName with advanced translation techniques. Another direction is accessing model performance on {\sc Indic-IndicXNLI} task,  where both premises and hypothesis are in two distinct \emph{Indic} languages.

\section*{Acknowledgement}
We thank members of the Utah NLP group for their valuable insights and suggestions at various stages of the project; and reviewers their helpful comments. We would also like to thank Suhani Aggarwal, Shibani Krishnatraya and Ayush Dhall for participating in the dataset verification activity and helping us find fluent speakers in many different indic languages. Additionally, we appreciate the inputs provided by Vivek Srikumar and Ellen Riloff. Vivek Gupta acknowledges support from Bloomberg’s Data Science Ph.D. Fellowship.

%% file: appendix.tex
\appendix

\section{Human Validation Scoring}
\label{sec:human-eval-guidelines-appendix}
We provide English and \emph{indic} language \datasetName (IndicTrans translated) sentence to the recruited native speaker of that \emph{indic} language for validation. Before the annotation work, each expert was given a full explanation of the guidelines that needed to be followed. The validation instructions (mturk template and detailed examples) are taken from the Semeval-2016 Task-I. The native speaker access the sentence pairs assign an integer score between \textbf{0} and \textbf{5}, as follows: \begin{inparaenum}[]
\item \textbf{0}: The two sentences are completely dissimilar. \item \textbf{1}: The two sentences are not equivalent, but are on the same topic. \item \textbf{2}: The two sentences are not equivalent, but share some details. \item \textbf{3}: The two sentences are roughly equivalent, but some important information differs/missing. \item \textbf{4}: The two sentences are mostly equivalent, but some unimportant details differ. \item \textbf{5}: The two sentences are exactly equivalent, as they mean the same.
\end{inparaenum} The score depicts the goodness of translated sentence in terms of semantics, i.e. same meaning as original English sentence\footnote{For NLI task, same syntax, i.e. grammar (e.g. Tense) lesser important than same Semantic, i.e. meaning preservation.}. Scores are then normalized to a probability range (between 0 and 1). The final validation score for each language is determined as the average of all 100 instances' scores. 

Additionally, we also computed the BERTScore between the English and the Hindi test split of the {\sc XNLI}\footnote{{\sc XNLI} hindi test splits was human translated.}, using multi-lingual strategy which came out to be 70 ($\times$ 10${^-2}$). We presume that the lower score is attributable to the fact that human-translated dataset encapsulates a large number of linguistic nuances, resulting in a change in the structure and tonality of the sentences, which is frequently overlooked by machine translation systems, as highlighted by \citet{bianchi2021language}.

\section{Details: Hyper Parameters Settings}
\label{sec:hyperparameters_appendix}
All the models were trained on google collaboratory \footnote{\url{https://colab.research.google.com/}} on TPU-v2 with 8 cores. The code was built in the PyTorch-lightning framework. We used accuracy as mentioned in the original XNLI paper \cite{conneau2018xnli} as our metric of choice. The training was run with an early stopping callback with the patience of 3, validation interval of 0.5 epochs and AdamW as optimizer\citep{loshchilov2019decoupled}.

\begin{table*}[htbp!]
\centering
\small
\begin{tabular}{l|rrrrrrrrr}\toprule
\textbf{Model} & \textbf{PO} & \textbf{CU}&\textbf{LR} & \textbf{BS} &\textbf{WD} &\textbf{MSL} &\textbf{MS} & \textbf{WS} \\
\midrule 
\textbf{XLM-R} & MLM (Dynamic) & Wikipedia Corpus &2e-5 &64 &0.01 &128 &278M &1500 \\
\textbf{iBERT} & MLM & IndicCorp &2e-5 &128 &0.01 &128 &33.7M &1500 \\
\textbf{MuRIL} & MLM, TLM and TrLM &OSCAR and PM India&2e-5 &64 &0.01 &128 &237M &1500 \\
\textbf{mBERT} & MLM & Wikipedia Corpus&2e-5 &128 &0.01 &128 &177M &1500 \\
\bottomrule
\end{tabular}
\caption{Model Hyper-Parameters}\label{tab: hyper-params}
\end{table*}

In Table \ref{tab: hyper-params} the hyperparamaters are abbreviated as mentioned below:
\begin{inparaenum}[(a.)]
    \item \textbf{PO:} Pre-training Objective. Where MLM stands for masked Language Modelling, TLM stands for Translation Language Modelling and TrLM stands for Transliteration Language Modelling,
    \item \textbf{CU:} Corpus Used,
    \item \textbf{LR:} Learning Rate,
    \item \textbf{BS:} Batch Size,
    \item \textbf{WD:} Weight Decay,
    \item \textbf{MSL:} Maximum Sequence Length,
    \item \textbf{MS:} Model Size described as number of parameters in millions,
    \item \textbf{WS}: Warm-up Step.
\end{inparaenum}

\section{Indic Cross-lingual Transfer}
\label{sec:indic_crosslingual_appendix}

\input{tables_partd}

This section is the extension of the \S \ref{sec:indic_crosslingual}. Table \ref{tab: Indic Cross-lingual transfer} are the cross-lingual transfer results of XLM-R, IndicBERT, mBERT and MuRIL respectively. The rows of the table consist of the languages on which the model is trained, while the columns represent the evaluation languages. E.g., in table \ref{tab: Indic Cross-lingual transfer} the first row represents that the model is trained on \emph{``as"} and then tested on all the languages in the column. The values in the row are the accuracy scores of the model when trained on the language in its leftmost column and tested on the language in its top-most row column.

\paragraph{XLM-R.}the model perform best for the \textit{``bn"} language. The model gives the best performance average across all other languages if trained on \textit{``bn"}. A model trained in other languages, on average, also performs best for \textit{``bn"} language. XLM-R also struggles to correlate with \textit{``kn"}, \textit{``or"}, and \textit{``ml"}, thus performs poorly on average if trained for them. At the same time, all models have poor cross-lingual ability transferability for the \textit{``as"} language. 

\paragraph{IndicBERT.}the overall score is comparable to XLM-R despite it's smaller size. On average, across languages, the cross-lingual transfer ability for models trained on varying \emph{indic} languages were consistently similar (b/w 0.5-0.6). However, the evaluation performance for cross-lingual models evaluated on \emph{``ml"} were poor for all \emph{indic} trained models. For model trained on some languages, \emph{``kn"}, \emph{``ml"} and \emph{``pa"}, the best performance was across diagonal, i.e. indicating the model performs best on the trained language. This trend was, however, was not shown in other \emph{indic} languages, indicating remarkable cross-lingual transfer ability of the IndicBERT model. 

\paragraph{mBERT.}the model performs worse for \textbf{``or"} on average for both when evaluated and train on. However, all models performs very consistently for other \textit{indic} languages. Model trained on \textit{kn}, \textit{pa}, \textit{ta}, \textit{hi}, and \textit{bn} perform best on average across languages. Here too, the best cross-lingual transfer ability was shown for \textit{bn} language. mBERT also have best performance across diagonal for some languages e.g. \textit{``as"}, \textit{``gu"}, \textit{``ml"}, \textit{``pa"} and \textit{``te"}.
\paragraph{MuRIL.} shows the best overall cross-lingual transfer ability amongst all the models. MuRIL only fails to generalize well when trained for \textit{``or"} language. However, model train on other \emph{indic} language when evaluated on \textit{``or"} performs well. Model trained on \textit{``ta"} and \textit{``ml"} performs best across all languages. The best cross-lingual transfer ability was shown for \textit{``bn"} and \textit{``hi"}. 
Overall, MuRIL has better cross-lingual transfer ability across all languages compared to other models. It also shows less performance bias for languages such as \textit{``bn"} and \textit{``hi"}, as compared to XLM-R.

%% file: tables_partd.tex
\begin{table*}[!htbp]\centering
\small
\begin{tabular}{m{3.5em}|m{0.3em}m{0.3em}m{0.3em}m{0.3em}m{0.3em}m{0.3em}m{0.3em}m{0.3em}m{0.3em}m{0.3em}m{0.7em}|m{3.2em}<{\centering} | m{0.3em}m{0.3em}m{0.3em}m{0.3em}m{0.3em}m{0.3em}m{0.3em}m{0.3em}m{0.3em}m{0.3em}m{0.7em}|m{3.3em}<{\centering}}
\toprule \\[-1.2em]
\multirow{2}{*}{\textbf{TrLang}} & \multicolumn{11}{c|}{\textbf{XLM-RoBERTa}} &  \multirow{2}{*}{\textbf{TrAvg}} & \multicolumn{11}{c|}{\textbf{IndicBERT}} &  \multirow{2}{*}{\textbf{TrAvg}} \\ \\[-1.2em]
\cmidrule(lr){2-12} \cmidrule(lr){14-24} \\[-1.2em]
& \textbf{as} & \textbf{gu} & \textbf{kn} &\textbf{ml} & \textbf{mr}& \textbf{or} & \textbf{pa} & \textbf{ta} & \textbf{te} & \textbf{bn} & \textbf{hi} & & \textbf{as} & \textbf{gu} & \textbf{kn} &\textbf{ml} & \textbf{mr}& \textbf{or} & \textbf{pa} & \textbf{ta} & \textbf{te} & \textbf{bn} & \textbf{hi} &\\\midrule 
\textbf{as} & 64 & 67 & 66 & 67 & 63 & 63 &\textbf{68}&\textbf{68} & 64 & 66 & 65 & 66 & 65 & 63 & 54 & 46 & 61 & 60 & 66 & 48 & 57 &\textbf{67} & 60 & 58\\
\textbf{gu} & 65 & 72 & 69 & 69 & 68 & 71 & 70 & 71 & 65 &\textbf{74} &\textbf{74} & 70 & 61 & 67 & 54 & 41 & 65 & 64 &\textbf{70} & 46 & 62 &\textbf{70} & 62 &\textbf{60}\\
\textbf{kn} & 33 & 31 &\textbf{35} &\textbf{35} & 31 & 34 & 32 & 31 & 32 & 33 & 32 & 33 & 58 & 64 &\textbf{68} & 48 & 59 & 59 & 65 & 46 & 59 & 63 & 63 & 59\\
\textbf{ml} &\textbf{35} & 33 & 33 & 34 & 31 & 34 & 34 & 31 & 33 & 34 & 34 & 33 & 55 & 52 & 54 &\textbf{60} & 53 & 53 & 52 & 52 & 57 & 52 & 52 & 54\\
\textbf{mr} & 66 & 74 & 70 & 72 & 72 & 68 & 70 & 69 & 65 &\textbf{75} & 73 & 71 & 62 & 65 & 54 & 48 & 61 & 61 & 67 & 52 & 60 &\textbf{68} & 63 &\textbf{60}\\
\textbf{or} & 35 & 33 & 32 & 36 & 35 & 34 & 34 &\textbf{36} & 34 &\textbf{36} &\textbf{36} & 35 & 61 & 66 & 57 & 49 & 61 & 66 & 65 & 48 & 60 &\textbf{68} & 64 &\textbf{60}\\
\textbf{pa} & 65 & 69 & 70 & 67 & 67 & 67 & 70 & 66 & 67 &\textbf{73} & 66 & 68 & 61 & 67 & 55 & 47 & 60 & 62 &\textbf{74} & 41 & 60 & 70 & 62 &\textbf{60} \\
\textbf{ta} & 64 & 67 & 69 & 72 & 71 & 68 & 71 & 70 & 70 &\textbf{73} & 70 & 70 & 55 &\textbf{60} & 53 & 49 & 56 & 54 & 58 & 59 & 55 & 58 & 55 & 56\\
\textbf{te} & 61 & 70 & 71 & 70 & 70 & 71 & 68 & 68 &\textbf{75} &\textbf{75} & 72 & 71 & 61 & 63 & 53 & 45 & 59 & 63 &\textbf{70} &46 & 63 & 68 & 58 & 59\\
\textbf{bn} & 67 & 72 & 73 & 73 & 72 &\textbf{74} &\textbf{74} & 70 & 70 & 73 & 71 &\textbf{72} & 62 & 66 & 55 & 48 & 62 & 62 & 66 & 47 & 60 &\textbf{68} &\textbf{68} &\textbf{60}\\
\textbf{hi} & 66 & 70 & 69 & 72 & 69 & 68 & 71 & 71 & 71 &\textbf{76} & 73 & 71 & 58 & 63 & 53 & 49 & 61 & 61 & 66 & 43 & 57 &\textbf{71} & 61 & 59 \\
\hline
\textbf{TestAvg} & 56 & 60 & 60 & 61 & 59 & 59 & 60 & 59 & 59 &\textbf{63} & 61 &60 & 60 & 63 & 55 & 48 & 60 & 60 & 65 & 48 & 59 &\textbf{66}& 61 & 59\\
\bottomrule
\multirow{2}{*}{\textbf{TrLang}} & \multicolumn{11}{c|}{\textbf{mBERT}} &  \multirow{2}{*}{\textbf{TrAvg}} & \multicolumn{11}{c|}{\textbf{MuRIL}} &  \multirow{2}{*}{\textbf{TrAvg}} \\ \\[-1.2em]
\cmidrule(lr){2-12} \cmidrule(lr){14-24} \\[-1.2em]
& \textbf{as} & \textbf{gu} & \textbf{kn} &\textbf{ml} & \textbf{mr}& \textbf{or} & \textbf{pa} & \textbf{ta} & \textbf{te} & \textbf{bn} & \textbf{hi} & & \textbf{as} & \textbf{gu} & \textbf{kn} &\textbf{ml} & \textbf{mr}& \textbf{or} & \textbf{pa} & \textbf{ta} & \textbf{te} & \textbf{bn} & \textbf{hi} &\\\midrule 
\textbf{as} &\textbf{69} & 59 & 61 & 53 & 57 & 36 & 61 & 57 & 52 & 59 & 64 & 56 & 73 &\textbf{78} &75 &74 &74 &73 &75 &75 &75 &76 &77 &75\\
\textbf{gu} & 48 &\textbf{70} & 64 & 55 & 60 & 32 & 64 & 64 & 60 & 67 & 65 & 60 &72 &75 &75 &74 &73 & 72 & 70 & 72 & 71 &\textbf{76} & 75 & 73\\
\textbf{kn} & 49 & 62 & 68 & 64 & 60 & 35 & 65 & 64 & 59 &\textbf{69} & 62 &\textbf{61} & 72 & 75 & 76 & 76 & 73 & 73 & 74 & 75 & 76 &\textbf{77} &\textbf{77} & 75\\
\textbf{ml} & 51 &60 & 60 &\textbf{71} & 60 & 30 & 61 & 65 & 62 & 66 & 62 & 60 & 75 & 75 & 73 & 77 & 72 & 78 & 76 &\textbf{79} & 75 & 77 & 76 &\textbf{76}\\
\textbf{mr} & 45 & 61 & 63 & 56 & 69 & 35 & 64 & 56 & 57 &\textbf{69} & 66 & 60 &69 &70 &72 &71 &\textbf{73} &68 &76 &70 &69 &73 &74 &72\\
\textbf{or} & 34 & 33 & 29 & 32 & 36 & 35 & 34 & 35 & 33 & 33 & 34 & 33 &33 &36 &35 &30 &32 &\textbf{35 }&30 &30 &33 &32 &36 &33 \\
\textbf{pa} & 47 & 65 & 59 & 59 & 62 & 35 &\textbf{70} & 63 & 61 & 68 & 64 &\textbf{61} &73 &75 &\textbf{76}&74 &74 &76 &79 &71 &74 &75 &75 &75\\
\textbf{ta} & 48 &64 &\textbf{67} &63 & 60 & 32 & 65 & 66 & 63 & 69 & 62 &\textbf{61} &74 &76 &76 &77 &75 &72 &74 &77 &76 &\textbf{80} &78 &\textbf{76} \\
\textbf{te} & 51 & 59 & 63 & 63 & 60 & 32 & 61 & 64 &\textbf{67} & 66 & 62 & 60 &70 &72 &74 &71 &73 &70 &\textbf{77}&74 &\textbf{77}&\textbf{77} &75 &74\\
\textbf{bn} & 51 & 64 & 65 & 62 & 62 & 32 & 65 & 60 & 62 & 69 &\textbf{67} &\textbf{61} &68 &\textbf{76} &73 &73 &71 &72 &73 &74 &74 &74 &\textbf{76} &74\\
\textbf{hi} & 50 & 66 & 65 & 61 & 62 & 30 & 65 & 63 & 61 &\textbf{71} & 63 &\textbf{61} &73 &76 &73 &75 &74 &73 &76 &74 &74 &75 &\textbf{76} &75\\
\hline
\textbf{Test Avg} & 49 & 60 & 60 & 58 & 59 & 33 & 61 & 60 & 58 &\textbf{64} & 61 & 58 & 68 & 71 &71 &70 &69 & 69 & 71 & 70 & 70 &\textbf{72} &\textbf{72}&71\\
\bottomrule
\end{tabular}
\caption{Indic Cross-lingual transfer}
\label{tab: Indic Cross-lingual transfer}
\centering
\small
\end{table*}

%% file: main.bbl
\begin{thebibliography}{49}
\expandafter\ifx\csname natexlab\endcsname\relax\def\natexlab#1{#1}\fi

\bibitem[{spe(2008)}]{spearman-correlation}
 2008.
\newblock \href {https://doi.org/10.1007/978-0-387-32833-1_379} {\emph{Spearman
  Rank Correlation Coefficient}}, pages 502--505. Springer New York, New York,
  NY.

\bibitem[{Aghajanyan et~al.(2021)Aghajanyan, Gupta, Shrivastava, Chen,
  Zettlemoyer, and Gupta}]{aghajanyan2021muppet}
Armen Aghajanyan, Anchit Gupta, Akshat Shrivastava, Xilun Chen, Luke
  Zettlemoyer, and Sonal Gupta. 2021.
\newblock \href {http://arxiv.org/abs/2101.11038} {Muppet: Massive multi-task
  representations with pre-finetuning}.

\bibitem[{Agirre et~al.(2016)Agirre, Banea, Cer, Diab, Gonzalez-Agirre,
  Mihalcea, Rigau, and Wiebe}]{agirre-etal-2016-semeval}
Eneko Agirre, Carmen Banea, Daniel Cer, Mona Diab, Aitor Gonzalez-Agirre, Rada
  Mihalcea, German Rigau, and Janyce Wiebe. 2016.
\newblock \href {https://doi.org/10.18653/v1/S16-1081} {{S}em{E}val-2016 task
  1: Semantic textual similarity, monolingual and cross-lingual evaluation}.
\newblock In \emph{Proceedings of the 10th International Workshop on Semantic
  Evaluation ({S}em{E}val-2016)}, pages 497--511, San Diego, California.
  Association for Computational Linguistics.

\bibitem[{Arora(2020)}]{arora-2020-inltk}
Gaurav Arora. 2020.
\newblock \href {https://doi.org/10.18653/v1/2020.nlposs-1.10} {i{NLTK}:
  Natural language toolkit for indic languages}.
\newblock In \emph{Proceedings of Second Workshop for NLP Open Source Software
  (NLP-OSS)}, pages 66--71, Online. Association for Computational Linguistics.

\bibitem[{Behr(2017)}]{doi:10.1080/13645579.2016.1252188}
Dorothée Behr. 2017.
\newblock \href {https://doi.org/10.1080/13645579.2016.1252188} {Assessing the
  use of back translation: the shortcomings of back translation as a quality
  testing method}.
\newblock \emph{International Journal of Social Research Methodology},
  20(6):573--584.

\bibitem[{Bianchi et~al.(2021)Bianchi, Nozza, and Hovy}]{bianchi2021language}
Federico Bianchi, Debora Nozza, and Dirk Hovy. 2021.
\newblock \href {http://arxiv.org/abs/2109.13037} {Language invariant
  properties in natural language processing}.

\bibitem[{Bowman et~al.(2015)Bowman, Angeli, Potts, and
  Manning}]{snli:emnlp2015}
Samuel~R. Bowman, Gabor Angeli, Christopher Potts, and Christopher~D. Manning.
  2015.
\newblock A large annotated corpus for learning natural language inference.
\newblock In \emph{Proceedings of the 2015 Conference on Empirical Methods in
  Natural Language Processing (EMNLP)}. Association for Computational
  Linguistics.

\bibitem[{Conneau et~al.(2020)Conneau, Khandelwal, Goyal, Chaudhary, Wenzek,
  Guzmán, Grave, Ott, Zettlemoyer, and Stoyanov}]{conneau2020unsupervised}
Alexis Conneau, Kartikay Khandelwal, Naman Goyal, Vishrav Chaudhary, Guillaume
  Wenzek, Francisco Guzmán, Edouard Grave, Myle Ott, Luke Zettlemoyer, and
  Veselin Stoyanov. 2020.
\newblock \href {http://arxiv.org/abs/1911.02116} {Unsupervised cross-lingual
  representation learning at scale}.

\bibitem[{Conneau and Lample(2019)}]{conneau2019cross}
Alexis Conneau and Guillaume Lample. 2019.
\newblock Cross-lingual language model pretraining.
\newblock \emph{Advances in Neural Information Processing Systems},
  32:7059--7069.

\bibitem[{Conneau et~al.(2018)Conneau, Rinott, Lample, Williams, Bowman,
  Schwenk, and Stoyanov}]{conneau2018xnli}
Alexis Conneau, Ruty Rinott, Guillaume Lample, Adina Williams, Samuel Bowman,
  Holger Schwenk, and Veselin Stoyanov. 2018.
\newblock \href {https://doi.org/10.18653/v1/D18-1269} {{XNLI}: Evaluating
  cross-lingual sentence representations}.
\newblock In \emph{Proceedings of the 2018 Conference on Empirical Methods in
  Natural Language Processing}, pages 2475--2485, Brussels, Belgium.
  Association for Computational Linguistics.

\bibitem[{Dabre et~al.(2021)Dabre, Shrotriya, Kunchukuttan, Puduppully, Khapra,
  and Kumar}]{dabre2021indicbart}
Raj Dabre, Himani Shrotriya, Anoop Kunchukuttan, Ratish Puduppully, Mitesh~M.
  Khapra, and Pratyush Kumar. 2021.
\newblock \href {http://arxiv.org/abs/2109.02903} {Indicbart: A pre-trained
  model for natural language generation of indic languages}.

\bibitem[{Dagan et~al.(2013)Dagan, Roth, Sammons, and
  Zanzotto}]{dagan2013recognizing}
Ido Dagan, Dan Roth, Mark Sammons, and Fabio~Massimo Zanzotto. 2013.
\newblock Recognizing textual entailment: Models and applications.
\newblock \emph{Synthesis Lectures on Human Language Technologies},
  6(4):1--220.

\bibitem[{Delbio et~al.(2018)Delbio, Abilasha, and m.~Ilankumaran}]{IJET23926}
A.~Delbio, R.~Abilasha, and m.~Ilankumaran. 2018.
\newblock \href {https://doi.org/10.14419/ijet.v7i4.36.23926} {Second language
  acquisition and mother tongue influence of english language learners – a
  psycho analytic approach}.
\newblock \emph{International Journal of Engineering and Technology},
  7(4.36):497--500.

\bibitem[{Devlin et~al.(2019)Devlin, Chang, Lee, and
  Toutanova}]{devlin2019bert}
Jacob Devlin, Ming-Wei Chang, Kenton Lee, and Kristina Toutanova. 2019.
\newblock \href {http://arxiv.org/abs/1810.04805} {Bert: Pre-training of deep
  bidirectional transformers for language understanding}.

\bibitem[{Dumitrescu et~al.(2021)Dumitrescu, Rebeja, Lorincz, Gaman, Avram,
  Ilie, Pruteanu, Stan, Rosia, Iacobescu, Morogan, Dima, Marchidan, Rebedea,
  Chitez, Yogatama, Ruder, Ionescu, Pascanu, and
  Patraucean}]{dumitrescu2021liro}
Stefan~Daniel Dumitrescu, Petru Rebeja, Beata Lorincz, Mihaela Gaman, Andrei
  Avram, Mihai Ilie, Andrei Pruteanu, Adriana Stan, Lorena Rosia, Cristina
  Iacobescu, Luciana Morogan, George Dima, Gabriel Marchidan, Traian Rebedea,
  Madalina Chitez, Dani Yogatama, Sebastian Ruder, Radu~Tudor Ionescu, Razvan
  Pascanu, and Viorica Patraucean. 2021.
\newblock \href {https://openreview.net/forum?id=JH61CD7afTv} {Liro: Benchmark
  and leaderboard for romanian language tasks}.
\newblock In \emph{Thirty-fifth Conference on Neural Information Processing
  Systems Datasets and Benchmarks Track (Round 1)}.

\bibitem[{Edunov et~al.(2020)Edunov, Ott, Ranzato, and
  Auli}]{edunov-etal-2020-evaluation}
Sergey Edunov, Myle Ott, Marc{'}Aurelio Ranzato, and Michael Auli. 2020.
\newblock \href {https://doi.org/10.18653/v1/2020.acl-main.253} {On the
  evaluation of machine translation systems trained with back-translation}.
\newblock In \emph{Proceedings of the 58th Annual Meeting of the Association
  for Computational Linguistics}, pages 2836--2846, Online. Association for
  Computational Linguistics.

\bibitem[{Goodfellow et~al.(2015)Goodfellow, Mirza, Xiao, Courville, and
  Bengio}]{goodfellow2015empirical}
Ian~J. Goodfellow, Mehdi Mirza, Da~Xiao, Aaron Courville, and Yoshua Bengio.
  2015.
\newblock \href {http://arxiv.org/abs/1312.6211} {An empirical investigation of
  catastrophic forgetting in gradient-based neural networks}.

\bibitem[{Graham et~al.(2020)Graham, Haddow, and
  Koehn}]{graham-etal-2020-statistical}
Yvette Graham, Barry Haddow, and Philipp Koehn. 2020.
\newblock \href {https://doi.org/10.18653/v1/2020.emnlp-main.6} {Statistical
  power and translationese in machine translation evaluation}.
\newblock In \emph{Proceedings of the 2020 Conference on Empirical Methods in
  Natural Language Processing (EMNLP)}, pages 72--81, Online. Association for
  Computational Linguistics.

\bibitem[{Haddow and Kirefu(2020)}]{haddow2020pmindia}
Barry Haddow and Faheem Kirefu. 2020.
\newblock \href {http://arxiv.org/abs/2001.09907} {Pmindia -- a collection of
  parallel corpora of languages of india}.

\bibitem[{Hu et~al.(2020)Hu, Ruder, Siddhant, Neubig, Firat, and
  Johnson}]{hu2020xtreme}
Junjie Hu, Sebastian Ruder, Aditya Siddhant, Graham Neubig, Orhan Firat, and
  Melvin Johnson. 2020.
\newblock \href {http://arxiv.org/abs/2003.11080} {Xtreme: A massively
  multilingual multi-task benchmark for evaluating cross-lingual
  generalization}.
\newblock \emph{CoRR}, abs/2003.11080.

\bibitem[{Jain et~al.(2020)Jain, Deshpande, Shridhar, Laumann, and
  Dash}]{jain2020indictransformers}
Kushal Jain, Adwait Deshpande, Kumar Shridhar, Felix Laumann, and Ayushman
  Dash. 2020.
\newblock \href {http://arxiv.org/abs/2011.02323} {Indic-transformers: An
  analysis of transformer language models for indian languages}.

\bibitem[{K et~al.(2021)K, Sathe, Aditya, and Choudhury}]{k2021analyzing}
Karthikeyan K, Aalok Sathe, Somak Aditya, and Monojit Choudhury. 2021.
\newblock \href
  {https://www.microsoft.com/en-us/research/publication/analyzing-the-effects-of-reasoning-types-on-cross-lingual-transfer-performance/}
  {Analyzing the effects of reasoning types on cross-lingual transfer
  performance}.
\newblock In \emph{EMNLP 2021}.

\bibitem[{Kakwani et~al.(2020)Kakwani, Kunchukuttan, Golla, N.C.,
  Bhattacharyya, Khapra, and Kumar}]{kakwani-etal-2020-indicnlpsuite}
Divyanshu Kakwani, Anoop Kunchukuttan, Satish Golla, Gokul N.C., Avik
  Bhattacharyya, Mitesh~M. Khapra, and Pratyush Kumar. 2020.
\newblock \href {https://doi.org/10.18653/v1/2020.findings-emnlp.445}
  {{I}ndic{NLPS}uite: Monolingual corpora, evaluation benchmarks and
  pre-trained multilingual language models for {I}ndian languages}.
\newblock In \emph{Findings of the Association for Computational Linguistics:
  EMNLP 2020}, pages 4948--4961, Online. Association for Computational
  Linguistics.

\bibitem[{Khanuja et~al.(2021)Khanuja, Bansal, Mehtani, Khosla, Dey, Gopalan,
  Margam, Aggarwal, Nagipogu, Dave, Gupta, Gali, Subramanian, and
  Talukdar}]{khanuja2021muril}
Simran Khanuja, Diksha Bansal, Sarvesh Mehtani, Savya Khosla, Atreyee Dey,
  Balaji Gopalan, Dilip~Kumar Margam, Pooja Aggarwal, Rajiv~Teja Nagipogu,
  Shachi Dave, Shruti Gupta, Subhash Chandra~Bose Gali, Vish Subramanian, and
  Partha Talukdar. 2021.
\newblock \href {http://arxiv.org/abs/2103.10730} {Muril: Multilingual
  representations for indian languages}.

\bibitem[{Kirch(2008)}]{pearson-correlation}
Wilhelm Kirch, editor. 2008.
\newblock \href {https://doi.org/10.1007/978-1-4020-5614-7_2569}
  {\emph{Pearson's Correlation Coefficient}}, pages 1090--1091. Springer
  Netherlands, Dordrecht.

\bibitem[{Kulesza(2012)}]{2012}
Alex Kulesza. 2012.
\newblock \href {https://doi.org/10.1561/2200000044} {Determinantal point
  processes for machine learning}.
\newblock \emph{Foundations and Trends® in Machine Learning},
  5(2-3):123–286.

\bibitem[{Kulesza and Taskar(2011)}]{10.5555/3104482.3104632}
Alex Kulesza and Ben Taskar. 2011.
\newblock K-dpps: Fixed-size determinantal point processes.
\newblock In \emph{Proceedings of the 28th International Conference on
  International Conference on Machine Learning}, ICML'11, page 1193–1200,
  Madison, WI, USA. Omnipress.

\bibitem[{Kunchukuttan(2020)}]{kunchukuttan2020indicnlp}
Anoop Kunchukuttan. 2020.
\newblock {The IndicNLP Library}.
\newblock
  \url{https://github.com/anoopkunchukuttan/indic_nlp_library/blob/master/docs/indicnlp.pdf}.

\bibitem[{Lee et~al.(2021)Lee, Vu, and Li}]{lee-etal-2021-meta}
Hung-yi Lee, Ngoc~Thang Vu, and Shang-Wen Li. 2021.
\newblock \href {https://doi.org/10.18653/v1/2021.acl-tutorials.3} {Meta
  learning and its applications to natural language processing}.
\newblock In \emph{Proceedings of the 59th Annual Meeting of the Association
  for Computational Linguistics and the 11th International Joint Conference on
  Natural Language Processing: Tutorial Abstracts}, pages 15--20, Online.
  Association for Computational Linguistics.

\bibitem[{Lewis et~al.(2020)Lewis, Oguz, Rinott, Riedel, and
  Schwenk}]{lewis-etal-2020-mlqa}
Patrick Lewis, Barlas Oguz, Ruty Rinott, Sebastian Riedel, and Holger Schwenk.
  2020.
\newblock \href {https://doi.org/10.18653/v1/2020.acl-main.653} {{MLQA}:
  Evaluating cross-lingual extractive question answering}.
\newblock In \emph{Proceedings of the 58th Annual Meeting of the Association
  for Computational Linguistics}, pages 7315--7330, Online. Association for
  Computational Linguistics.

\bibitem[{Liang et~al.(2020)Liang, Duan, Gong, Wu, Guo, Qi, Gong, Shou, Jiang,
  Cao, Fan, Zhang, Agrawal, Cui, Wei, Bharti, Qiao, Chen, Wu, Liu, Yang,
  Campos, Majumder, and Zhou}]{liang-etal-2020-xglue}
Yaobo Liang, Nan Duan, Yeyun Gong, Ning Wu, Fenfei Guo, Weizhen Qi, Ming Gong,
  Linjun Shou, Daxin Jiang, Guihong Cao, Xiaodong Fan, Ruofei Zhang, Rahul
  Agrawal, Edward Cui, Sining Wei, Taroon Bharti, Ying Qiao, Jiun-Hung Chen,
  Winnie Wu, Shuguang Liu, Fan Yang, Daniel Campos, Rangan Majumder, and Ming
  Zhou. 2020.
\newblock \href {https://doi.org/10.18653/v1/2020.emnlp-main.484} {{XGLUE}: A
  new benchmark datasetfor cross-lingual pre-training, understanding and
  generation}.
\newblock In \emph{Proceedings of the 2020 Conference on Empirical Methods in
  Natural Language Processing (EMNLP)}, pages 6008--6018, Online. Association
  for Computational Linguistics.

\bibitem[{Liu et~al.(2020)Liu, Gu, Goyal, Li, Edunov, Ghazvininejad, Lewis, and
  Zettlemoyer}]{liu-etal-2020-multilingual-denoising}
Yinhan Liu, Jiatao Gu, Naman Goyal, Xian Li, Sergey Edunov, Marjan
  Ghazvininejad, Mike Lewis, and Luke Zettlemoyer. 2020.
\newblock \href {https://doi.org/10.1162/tacl_a_00343} {Multilingual denoising
  pre-training for neural machine translation}.
\newblock \emph{Transactions of the Association for Computational Linguistics},
  8:726--742.

\bibitem[{Loshchilov and Hutter(2019)}]{loshchilov2019decoupled}
Ilya Loshchilov and Frank Hutter. 2019.
\newblock \href {http://arxiv.org/abs/1711.05101} {Decoupled weight decay
  regularization}.

\bibitem[{Miyabe and Yoshino(2015)}]{7433246}
Mai Miyabe and Takashi Yoshino. 2015.
\newblock \href {https://doi.org/10.1109/Culture.and.Computing.2015.35}
  {Evaluation of the validity of back-translation as a method of assessing the
  accuracy of machine translation}.
\newblock In \emph{2015 International Conference on Culture and Computing
  (Culture Computing)}, pages 145--150.

\bibitem[{Ortiz~Su{\'a}rez et~al.(2020)Ortiz~Su{\'a}rez, Romary, and
  Sagot}]{ortiz-suarez-etal-2020-monolingual}
Pedro~Javier Ortiz~Su{\'a}rez, Laurent Romary, and Beno{\^\i}t Sagot. 2020.
\newblock \href {https://doi.org/10.18653/v1/2020.acl-main.156} {A monolingual
  approach to contextualized word embeddings for mid-resource languages}.
\newblock In \emph{Proceedings of the 58th Annual Meeting of the Association
  for Computational Linguistics}, pages 1703--1714, Online. Association for
  Computational Linguistics.

\bibitem[{{Ortiz Su{'a}rez} et~al.(2019){Ortiz Su{'a}rez}, Sagot, and
  Romary}]{OrtizSuarezSagotRomary2019}
Pedro~Javier {Ortiz Su{'a}rez}, Benoit Sagot, and Laurent Romary. 2019.
\newblock \href {https://doi.org/10.14618/ids-pub-9021} {Asynchronous pipelines
  for processing huge corpora on medium to low resource infrastructures}.
\newblock Proceedings of the Workshop on Challenges in the Management of Large
  Corpora (CMLC-7) 2019. Cardiff, 22nd July 2019, pages 9 -- 16, Mannheim.
  Leibniz-Institut f{"u}r Deutsche Sprache.

\bibitem[{Papineni et~al.(2002)Papineni, Roukos, Ward, and
  Zhu}]{10.3115/1073083.1073135}
Kishore Papineni, Salim Roukos, Todd Ward, and Wei-Jing Zhu. 2002.
\newblock \href {https://doi.org/10.3115/1073083.1073135} {Bleu: A method for
  automatic evaluation of machine translation}.
\newblock In \emph{Proceedings of the 40th Annual Meeting on Association for
  Computational Linguistics}, ACL '02, page 311–318, USA. Association for
  Computational Linguistics.

\bibitem[{Pires et~al.(2019)Pires, Schlinger, and
  Garrette}]{pires-etal-2019-multilingual}
Telmo Pires, Eva Schlinger, and Dan Garrette. 2019.
\newblock \href {https://doi.org/10.18653/v1/P19-1493} {How multilingual is
  multilingual {BERT}?}
\newblock In \emph{Proceedings of the 57th Annual Meeting of the Association
  for Computational Linguistics}, pages 4996--5001, Florence, Italy.
  Association for Computational Linguistics.

\bibitem[{Radford and Narasimhan(2018)}]{Radford2018ImprovingLU}
Alec Radford and Karthik Narasimhan. 2018.
\newblock Improving language understanding by generative pre-training.

\bibitem[{Ramesh et~al.(2021)Ramesh, Doddapaneni, Bheemaraj, Jobanputra, AK,
  Sharma, Sahoo, Diddee, J, Kakwani, Kumar, Pradeep, Deepak, Raghavan,
  Kunchukuttan, Kumar, and Khapra}]{ramesh2021samanantar}
Gowtham Ramesh, Sumanth Doddapaneni, Aravinth Bheemaraj, Mayank Jobanputra,
  Raghavan AK, Ajitesh Sharma, Sujit Sahoo, Harshita Diddee, Mahalakshmi J,
  Divyanshu Kakwani, Navneet Kumar, Aswin Pradeep, Kumar Deepak, Vivek
  Raghavan, Anoop Kunchukuttan, Pratyush Kumar, and Mitesh~Shantadevi Khapra.
  2021.
\newblock \href {http://arxiv.org/abs/2104.05596} {Samanantar: The largest
  publicly available parallel corpora collection for 11 indic languages}.

\bibitem[{Rapp(2009)}]{10.5555/1667583.1667625}
Reinhard Rapp. 2009.
\newblock \href {https://aclanthology.org/P09-2034} {The back-translation
  score: Automatic mt evaluation at the sentence level without reference
  translations}.
\newblock In \emph{Proceedings of the ACL-IJCNLP 2009 Conference Short Papers},
  ACLShort '09, page 133–136, USA. Association for Computational Linguistics.

\bibitem[{Ruder et~al.(2021)Ruder, Constant, Botha, Siddhant, Firat, Fu, Liu,
  Hu, Garrette, Neubig, and Johnson}]{ruder-etal-2021-xtreme}
Sebastian Ruder, Noah Constant, Jan Botha, Aditya Siddhant, Orhan Firat, Jinlan
  Fu, Pengfei Liu, Junjie Hu, Dan Garrette, Graham Neubig, and Melvin Johnson.
  2021.
\newblock \href {https://aclanthology.org/2021.emnlp-main.802} {{XTREME}-{R}:
  Towards more challenging and nuanced multilingual evaluation}.
\newblock In \emph{Proceedings of the 2021 Conference on Empirical Methods in
  Natural Language Processing}, pages 10215--10245, Online and Punta Cana,
  Dominican Republic. Association for Computational Linguistics.

\bibitem[{Uppal et~al.(2020)Uppal, Gupta, Swaminathan, Zhang, Mahata, Gosangi,
  Shah, and Stent}]{uppal-etal-2020-two}
Shagun Uppal, Vivek Gupta, Avinash Swaminathan, Haimin Zhang, Debanjan Mahata,
  Rakesh Gosangi, Rajiv~Ratn Shah, and Amanda Stent. 2020.
\newblock \href {https://aclanthology.org/2020.aacl-main.71} {Two-step
  classification using recasted data for low resource settings}.
\newblock In \emph{Proceedings of the 1st Conference of the Asia-Pacific
  Chapter of the Association for Computational Linguistics and the 10th
  International Joint Conference on Natural Language Processing}, pages
  706--719, Suzhou, China. Association for Computational Linguistics.

\bibitem[{Vaswani et~al.(2017)Vaswani, Shazeer, Parmar, Uszkoreit, Jones,
  Gomez, Kaiser, and Polosukhin}]{vaswani2017attention}
Ashish Vaswani, Noam Shazeer, Niki Parmar, Jakob Uszkoreit, Llion Jones,
  Aidan~N Gomez, {\L}ukasz Kaiser, and Illia Polosukhin. 2017.
\newblock Attention is all you need.
\newblock In \emph{Advances in neural information processing systems}, pages
  5998--6008.

\bibitem[{Wenzek et~al.(2020)Wenzek, Lachaux, Conneau, Chaudhary, Guzm{\'a}n,
  Joulin, and Grave}]{wenzek-etal-2020-ccnet}
Guillaume Wenzek, Marie-Anne Lachaux, Alexis Conneau, Vishrav Chaudhary,
  Francisco Guzm{\'a}n, Armand Joulin, and Edouard Grave. 2020.
\newblock \href {https://aclanthology.org/2020.lrec-1.494} {{CCN}et: Extracting
  high quality monolingual datasets from web crawl data}.
\newblock In \emph{Proceedings of the 12th Language Resources and Evaluation
  Conference}, pages 4003--4012, Marseille, France. European Language Resources
  Association.

\bibitem[{Williams et~al.(2018)Williams, Nangia, and Bowman}]{N18-1101}
Adina Williams, Nikita Nangia, and Samuel Bowman. 2018.
\newblock \href {http://aclweb.org/anthology/N18-1101} {A broad-coverage
  challenge corpus for sentence understanding through inference}.
\newblock In \emph{Proceedings of the 2018 Conference of the North American
  Chapter of the Association for Computational Linguistics: Human Language
  Technologies, Volume 1 (Long Papers)}, pages 1112--1122. Association for
  Computational Linguistics.

\bibitem[{Xue et~al.(2021)Xue, Constant, Roberts, Kale, Al-Rfou, Siddhant,
  Barua, and Raffel}]{xue-etal-2021-mt5}
Linting Xue, Noah Constant, Adam Roberts, Mihir Kale, Rami Al-Rfou, Aditya
  Siddhant, Aditya Barua, and Colin Raffel. 2021.
\newblock \href {https://doi.org/10.18653/v1/2021.naacl-main.41} {m{T}5: A
  massively multilingual pre-trained text-to-text transformer}.
\newblock In \emph{Proceedings of the 2021 Conference of the North American
  Chapter of the Association for Computational Linguistics: Human Language
  Technologies}, pages 483--498, Online. Association for Computational
  Linguistics.

\bibitem[{Yang et~al.(2019)Yang, Zhang, Tar, and
  Baldridge}]{yang-etal-2019-paws}
Yinfei Yang, Yuan Zhang, Chris Tar, and Jason Baldridge. 2019.
\newblock \href {https://doi.org/10.18653/v1/D19-1382} {{PAWS}-{X}: A
  cross-lingual adversarial dataset for paraphrase identification}.
\newblock In \emph{Proceedings of the 2019 Conference on Empirical Methods in
  Natural Language Processing and the 9th International Joint Conference on
  Natural Language Processing (EMNLP-IJCNLP)}, pages 3687--3692, Hong Kong,
  China. Association for Computational Linguistics.

\bibitem[{Zhang* et~al.(2020)Zhang*, Kishore*, Wu*, Weinberger, and
  Artzi}]{bert-score}
Tianyi Zhang*, Varsha Kishore*, Felix Wu*, Kilian~Q. Weinberger, and Yoav
  Artzi. 2020.
\newblock \href {https://openreview.net/forum?id=SkeHuCVFDr} {Bertscore:
  Evaluating text generation with bert}.
\newblock In \emph{International Conference on Learning Representations}.

\end{thebibliography}
